\documentclass{format_arxiv/bytedance}
\newcommand{\arxivVersion}{} 
\usepackage{amsmath,amsfonts,amsthm,amssymb}
\usepackage{mathtools}
\usepackage{bm}
\usepackage{nicefrac}
\usepackage{animate}

\usepackage{xcolor}
\usepackage{epsfig}
\usepackage{graphicx}

\usepackage{adjustbox}
\usepackage{array}
\newcolumntype{C}[1]{>{\centering\arraybackslash}p{#1}}
\usepackage{booktabs}
\usepackage{colortbl}
\usepackage{wrapfig}
\usepackage{hhline}
\usepackage{multirow}
\usepackage{subcaption}

\usepackage{changepage}
\usepackage{extramarks}
\usepackage{fancyhdr}
\usepackage{lastpage}
\usepackage{setspace}
\usepackage{soul}
\usepackage{xspace}
\usepackage{adjustbox}

\usepackage{algpseudocode}
\usepackage{algorithmicx}
\usepackage[ruled]{algorithm2e}
\usepackage{enumitem}  
\usepackage{makecell}
\usepackage{pifont}
\usepackage{fontawesome}
\usepackage{ifthen}
\usepackage{etoolbox}
\usepackage{bbding}
\usepackage{pifont}
\usepackage{multirow}

\usepackage{listings}
\usepackage{xcolor}

\usepackage{float}

\newcommand{\ignore}[1]{}

\DeclareMathAlphabet{\mathbfit}{OML}{cmm}{b}{it}

\makeatletter
\DeclareRobustCommand\onedot{\futurelet\@let@token\@onedot}
\def\@onedot{\ifx\@let@token.\else.\null\fi\xspace}

\makeatother

\definecolor{MyDarkBlue}{rgb}{0,0.08,1}
\definecolor{MyAqua}{rgb}{0,0.7,0.7}
\definecolor{MyDarkGreen}{rgb}{0.02,0.6,0.02}
\definecolor{MyDarkOrange}{rgb}{0.40,0.2,0.02}
\definecolor{MyPurple}{HTML}{7338a0}
\definecolor{MyPink}{RGB}{250, 147, 240}
\definecolor{MyRed}{RGB}{227, 44, 94}
\definecolor{MyBlue}{RGB}{69, 144, 201}
\definecolor{MyOrange}{HTML}{d6970e}
\definecolor{MyPink}{HTML}{D86ECC}
\definecolor{MyGold}{rgb}{0.75,0.6,0.12}
\definecolor{MyDarkgray}{rgb}{0.66, 0.66, 0.66}

\def\enableReplayAnimation{}

\newcommand{\IfDefinedSwitch}[3]{%
  \ifdefined#1%
    #2%
  \else%
    #3%
  \fi%
}

\newcolumntype{M}[1]{>{\centering\arraybackslash}m{#1}}

\DeclareFontShape{T1}{bytesans}{b}{n}{<-> s * [1] format_arxiv/bytesans}{}
\DeclareFontShape{T1}{bytesans}{bx}{n}{<-> s * [1] format_arxiv/bytesans}{}
\DeclareFontShape{T1}{bytesans}{m}{sc}{<-> s * [1] format_arxiv/bytesans}{}
\DeclareFontShape{T1}{bytesans}{b}{sc}{<-> s * [1] format_arxiv/bytesans}{}
\DeclareFontShape{T1}{bytesans}{bx}{sc}{<-> s * [1] format_arxiv/bytesans}{}
\let\arxivincludegraphics\includegraphics
\renewcommand{\includegraphics}[2][]{%
  \ifstrequal{#2}{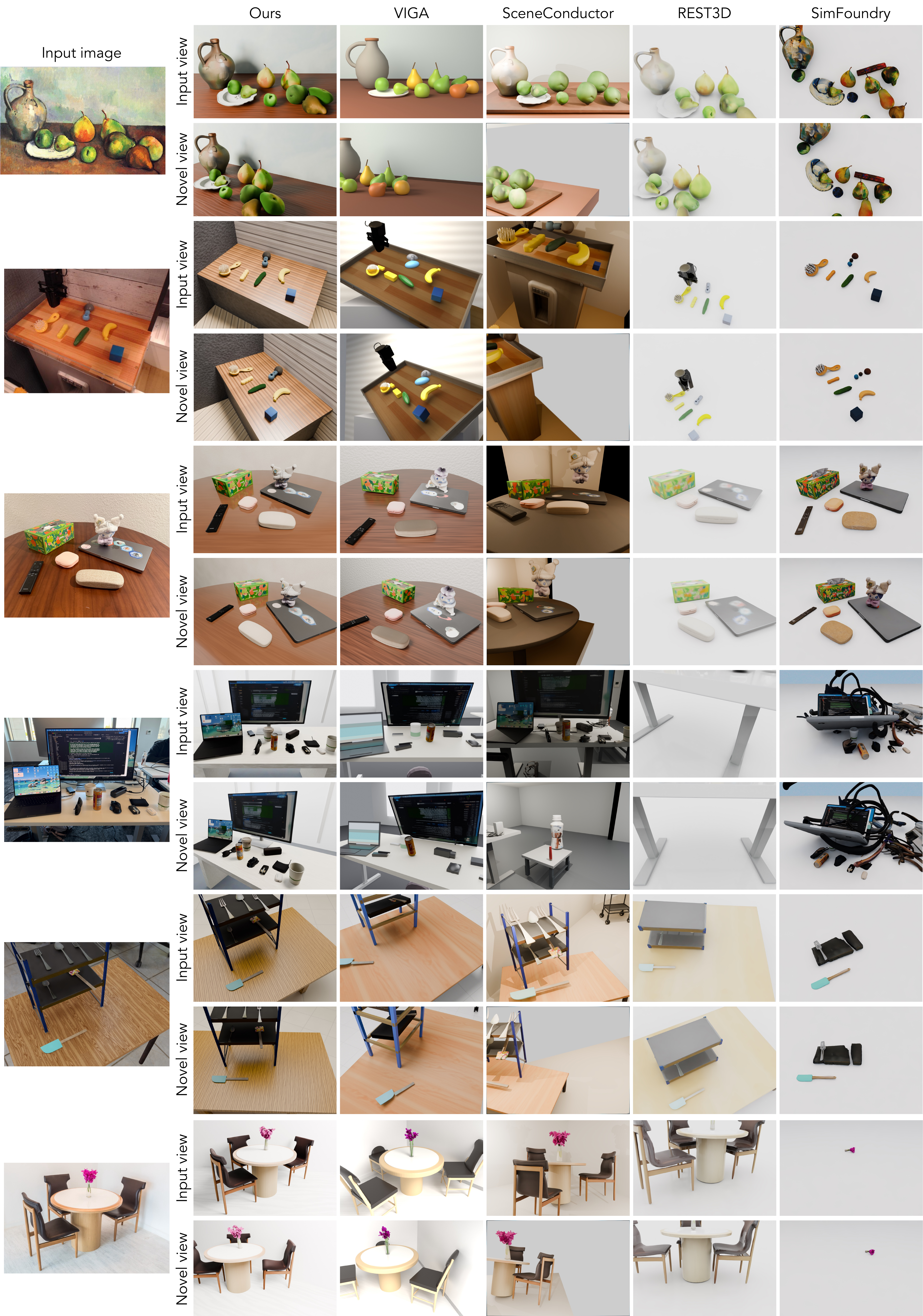}%
    {\arxivincludegraphics[width=\linewidth,height=0.88\textheight,keepaspectratio]{#2}}%
    {\arxivincludegraphics[#1]{#2}}%
}

\fancypagestyle{firststyle}{%
  \fancyhf{}%
  \fancyhead[R]{\raisebox{-10mm}[0pt][0pt]{\includegraphics[height=5mm]{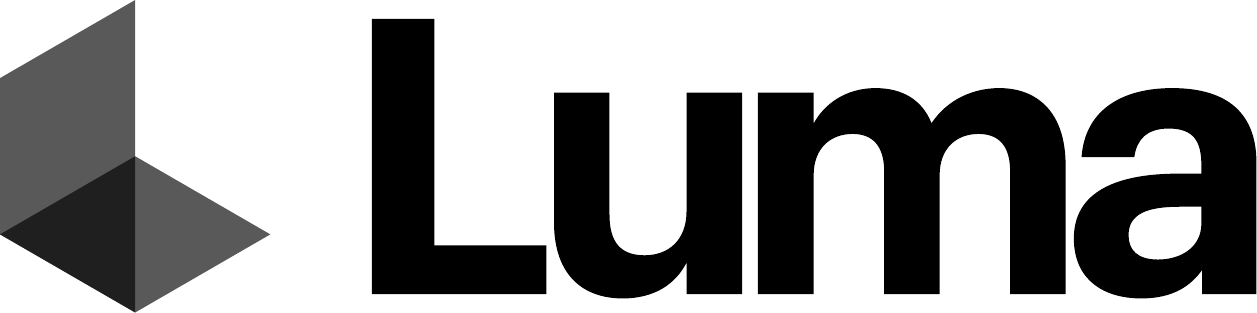}}}%
  \fancyfoot[C]{\thepage}%
}

\definecolor{seedbg}{HTML}{8C8C8C}
\definecolor{seedblue}{HTML}{000000}
\renewcommand{\seedblue}[1]{{\bfseries\color{seedblue}#1}}

\makeatletter
\renewcommand{\@toptitlebar}{\color{seedbg}\hrule height 0.5pt\color{black}\vskip 6mm}
\renewcommand{\@bottomtitlebar}{\vskip 6mm\color{seedbg}\hrule height 0.5pt\color{black}}
\makeatother

\newcommand{\method}{\textsc{SceneRig}\xspace}
\newcommand{\arxivCodeURL}{https://github.com/Red-Fairy/SceneRig}
\newcommand{\arxivProjectURL}{https://red-fairy.github.io/SceneRig/}
\newcommand{\anonymousProjectURL}{https://scenerig.github.io/}
\newcommand{\arxivSwitch}[2]{\IfDefinedSwitch{\arxivVersion}{#1}{#2}}
\newcommand{\page}{\arxivSwitch
  {\href{\arxivProjectURL}{project page}}
  {\href{\anonymousProjectURL}{project page}}}

\ifdefined\arxivVersion
  \RenewDocumentCommand{\vspace}{s m}{}%
\fi

\title{\method: An Agentic System for Simulation-Ready 3D Scene Reconstruction from Single Images}
\author[1,2,*]{Rundong Luo}
\author[1,\dagger]{Yunong Liu}
\author[1,\dagger]{Di Cao}
\author[1]{\\Matthew Tancik}
\author[1]{Shyamal Buch}
\author[1]{Colton Stearns}
\author[1]{Wenqi Xian}
\affiliation[1]{Luma AI}
\affiliation[2]{Cornell University}
\checkdata[Project Page]{\arxivSwitch
  {\expandafter\url\expandafter{\arxivProjectURL}}
  {\expandafter\url\expandafter{\anonymousProjectURL}}}
\begingroup
\RenewDocumentEnvironment{figure}{O{} +b}{\gdef\arxivteaser{\begin{figure}[t]#2\end{figure}}}{}
\RenewDocumentEnvironment{abstract}{+b}{\gdef\abstractlist{{\abstractinfont #1}}}{}
\begin{figure}[h]
    \centering
    \vspace{-8mm}
    \includegraphics[width=\linewidth]{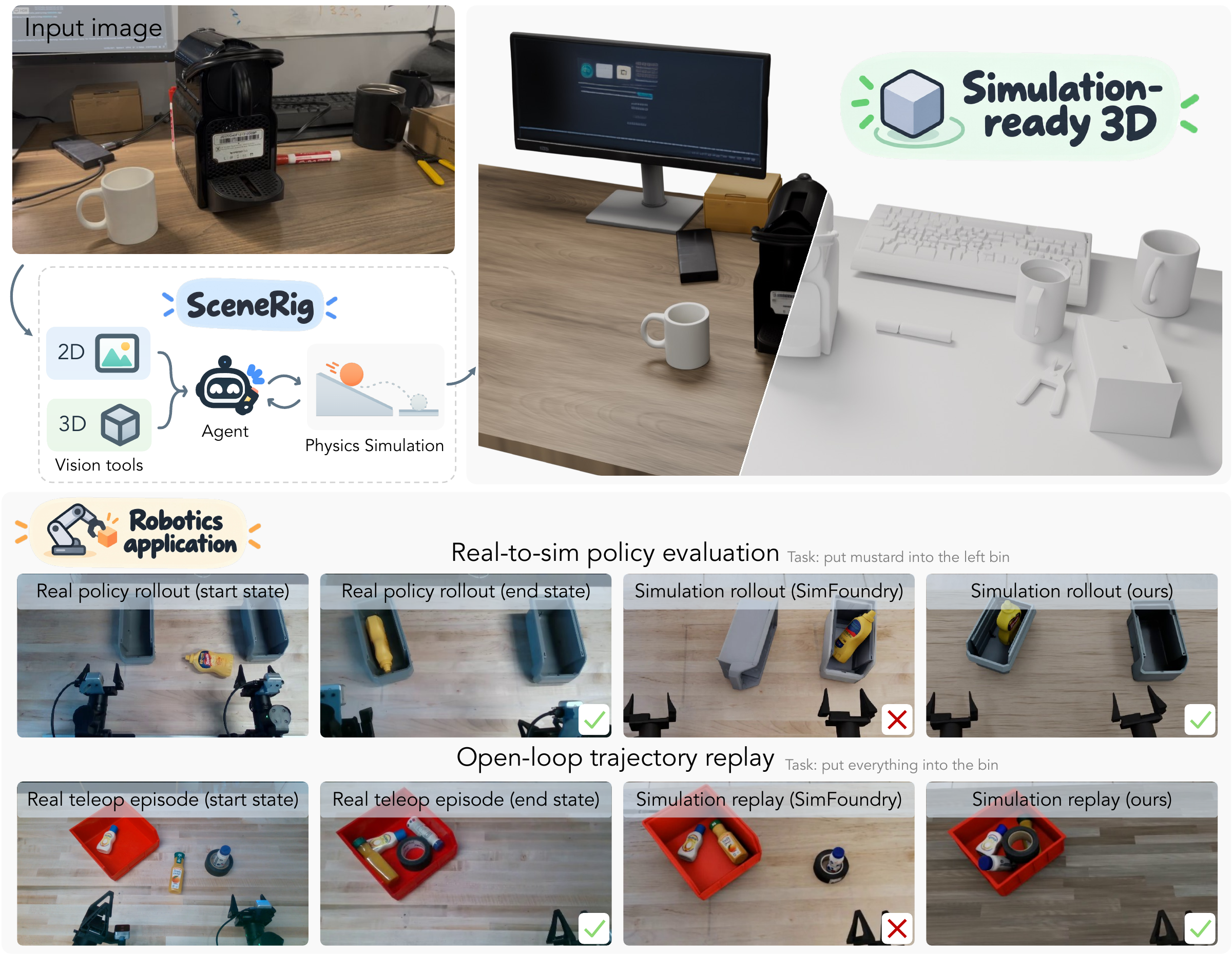}
    \vspace{-6mm}
    \caption{\textbf{Single-image simulation-ready scene reconstruction for robotics.} \method combines vision tools, agentic reasoning, and physical simulation to create an editable, simulation-ready scene from a single image (top). Its reconstructions support real-to-sim policy evaluation and open-loop trajectory replay, reproducing real-world outcomes more faithfully than SimFoundry (bottom).}
    \label{fig:teaser}
    \vspace{-1mm}
\end{figure}

\begin{abstract}

Simulation is a powerful tool for robotics, but building diverse, simulation-ready scenes at scale remains difficult. Reconstructing scenes from single images is a promising alternative, but visual similarity alone is insufficient: objects must also be recovered, accurately placed, and physically grounded. Existing methods that rely on post-hoc physics settling often misplace objects, producing scenes that look correct but behave poorly under manipulation.
We present \method, an agentic system that turns a single image into an editable, simulation-ready scene.  \method combines geometric, visual, and physical tools to reconstruct a faithful 3D scene. An agent then performs edit-and-validate loops inside a physics simulator to achieve physical stability.
Across diverse inputs, \method improves instance segmentation, object-centric appearance, and 3D placement while keeping the reconstructed scenes physically stable. Furthermore, \method is effective for downstream robotics: its scenes predict real-robot policy outcomes more accurately than the baseline method and more reliably support replay of recorded manipulations without trajectory adaptation.
\arxivSwitch{}{Link to the anonymous~\page.}

\end{abstract}

\endgroup

\makeatletter
\providecommand{\@LN}[2]{}
\providecommand{\@LN@col}[1]{}
\makeatother

\begin{document}
\maketitle
\footnotetext[1]{Work done during an internship at Luma AI.}
\footnotetext[2]{Equal contribution.}
\setcounter{footnote}{2}
\arxivteaser
\section{Introduction}
\label{sec:intro}

Evaluating the quality and robustness of action policies remains a fundamental bottleneck in robotics.
While physical execution is the gold standard, on-robot evaluations are notoriously difficult to scale: they demand expensive hardware, human supervision, and repeated trials across tasks and conditions~\citep{li2024simpler}.
Consequently, simulation is a powerful alternative that enables large-scale policy evaluation~\citep{yang2026robolab}.
However, scaling these environments remains difficult, as constructing realistic scenes requires substantial manual effort~\citep{jain2025polaris}.
%

Recent advances in agentic reconstruction and 3D generation have made real-to-sim scene reconstruction a promising route to scaling simulation environments~\citep{viga2026,chen2026sam3d,simfoundry2026}.
In fact, real-to-sim digital twins go beyond mere scaling: they capture the objects and arrangements of physical workspaces and let anyone turn a single photo into a simulation environment for policy evaluation.
Yet, despite recent progress in real-to-sim, balancing high-quality 3D reconstruction with faithful, robot-ready simulation physics remains challenging~\citep{ma2026rest3d,chen2026agentic}.
Even when a reconstruction is physically settled, objects can move away from their observed positions, producing a stable scene that no longer matches the robot's workspace.
To bridge this gap, we propose \method, an agentic system for simulation-ready 3D scene reconstruction from a single image.
\method addresses this problem by coupling image-guided pose refinement with physical validation for faithful trajectory replay and policy evaluation.

Making this coupling effective requires the agent to inspect the effects of physical simulation and use that feedback to guide further corrections.
%
Even GPT-6 Astra struggles to achieve accurate, physically stable object placement through VIGA or a general-purpose coding agent when the reconstructed scenes are evaluated after physical settling (Table~\ref{tab:astra-harness}).
%
\method's core contribution is therefore a harness that connects the agent’s visual reasoning with geometric and physical tools for accurate, stable scene reconstruction. The agent uses the updated scene and feedback from the edit to decide what to investigate or correct next.
%

More concretely, \method uses a flexible, modular harness with four stages.
First, \method builds a \textit{relational scene graph}, identifying objects and their support relationships.
Next, \textit{scene object creation} uses 2D and 3D vision 
models to estimate metric geometry and reconstruct object meshes. We then assign physical properties and initialize object placement through geometric alignment and support-ordered settling.
Third, specialized agents construct the surrounding geometry, materials, and lighting, with
strict rule-based verification gates for geometry and visual verification of all three.
Finally, and most importantly, \method executes a \textit{physical edit-and-validate loop}.
The agent selects which object to adjust and whether to change its position, scale, or orientation; geometric and physical tools search for an adjustment and test its effects in simulation.
%
Although the current implementation leverages specific state-of-the-art models, \method is designed to incorporate future foundation models.

We evaluate \method on diverse images, spanning synthetic renderings, real robot workspaces, AI-generated images, and in-the-wild smartphone captures.
We demonstrate that SceneRig improves instance segmentation, object-centric appearance, and 3D object placement over existing methods, with all evaluated scenes passing physical stability checks (Tables~\ref{tab:vlm}, \ref{tab:object-centric}).
Beyond geometric reconstruction, we show that \method yields robot-ready environments (Figure~\ref{fig:teaser}), supporting both real-to-sim policy evaluation and open-loop trajectory replay.
%
In summary, our contributions are:   \begin{list}{\textbullet}{%
\settowidth{\labelwidth}{\textbullet}%
\setlength{\labelsep}{0.5em}%
\setlength{\leftmargin}{\labelwidth}%
\addtolength{\leftmargin}{\labelsep}%
\setlength{\topsep}{0pt}%
\setlength{\partopsep}{0pt}%
\setlength{\itemsep}{2pt}%
\setlength{\parsep}{0pt}%
}
\item We introduce \method, an agentic system that reconstructs simulation-ready digital twins from a single image, prioritizing physical grounding in addition to visual similarity.
\item We build a modular harness that equips agents with geometric and physical tools. 
Through iterative edit-and-validate loops, \method directly incorporates physics into pose refinement.
\item \method improves reconstruction quality and robotics performance over single-image real-to-sim systems. Compared with the baseline, it increases open-loop trajectory replay success from 36\% to 80\% and episode-level agreement with real-world policy result from 54\% to 69\%.



\end{list}  
\section{Related Work}
\label{sec:related}

\noindent\textbf{Single-Image Scene Reconstruction.}
Object-centric reconstruction recovers individual objects and their spatial arrangement. Supervised methods predict geometry or retrieve assets: Total3D~\citep{nie2020total3d} predicts indoor layout and meshes, while ROCA~\citep{gumerli2022roca} retrieves and aligns CAD models. Without object annotations, uORF~\citep{yu2022uorf} and uOCF~\citep{luo2024uocf} learn compositional neural fields from multi-view images to separate objects and backgrounds. For explicit meshes, generative methods reconstruct instances separately and align them~\citep{dogaru2025gen3dsr,zhou2024dpa,han2025reparo,zhao2025depr}, while MIDI~\citep{huang2025midi} jointly generates multiple objects. CAST~\citep{yao2025cast} combines occlusion-completed object generation with pose refinement using scene relations and signed-distance constraints. SAM3D~\citep{chen2026sam3d} predicts object geometry, texture, and camera-relative layout. However, visual fidelity alone does not ensure physical stability or readiness for simulation.

\noindent\textbf{Agentic 3D Scene Creation.}
VLM agents construct 3D scenes through tool use and iterative feedback~\citep{hu2024scenecraft,huang2024blenderalchemy}. To ground generation and editing in scene structure, \citet{liu2025agentic3d} maintain labeled point clouds and scene relations. SceneWeaver~\citep{yang2025sceneweaver} adaptively selects synthesis and editing tools, while Scenethesis~\citep{ling2025scenethesis} combines language-based planning with generated-image guidance and geometric constraints. Beyond planning and tool selection, SAGE~\citep{xia2026sage} and SceneSmith~\citep{pfaff2026scenesmith} use editor-critic loops for indoor scene generation. 
For precise placement, agents also use numerical solvers~\citep{yang2024holodeck,sun2025layoutvlm,wang2026physcensis}.

Agentic workflows also support reconstruction from visual references. VIGA~\citep{viga2026} reconstructs 3D scenes with a generator--verifier loop, while SEIG~\citep{seig2026} extends this approach to refine geometry, materials, and lighting sequentially. SceneConductor~\citep{kim2026sceneconductor} initializes layouts, builds environments from point maps, and delegates local corrections to specialist agents. For evaluation, BlenderGym~\citep{gu2025blendergym} benchmarks graphics editing, while IR3D-Bench~\citep{liu2025ir3dbench} evaluates scene understanding through agentic inverse rendering. In contrast, \method's agent corrects objects using geometry- and physics-grounded tools.

\noindent\textbf{Physically Grounded Real-to-Sim.}
Real-to-sim aims at building interactive environments for robot learning and evaluation. RialTo~\citep{torne2024rialto}, Re3Sim~\citep{han2025re3sim}, and HoloScene~\citep{xia2025holoscene} recover interactive assets from video or scans. Agentic Real2Sim~\citep{chen2026agentic} turns a recorded robot--object interaction into a simulatable twin. Our closest baseline, SimFoundry~\citep{simfoundry2026}, reconstructs editable digital twins from video and generates digital cousins for real-to-sim policy evaluation and sim-to-real policy training. For data generation without dynamics simulation, Real2Render2Real~\citep{yu2025real2render2real} renders robot training images from scans. 

Concurrent single-image methods use several correction strategies. TableVerse~\citep{tableverse2026} resolves collisions through geometric translations before settling; $\phi$-Scene~\citep{li2026phiscene} optimizes clearance and settles objects in support order. REST3D~\citep{ma2026rest3d} optimizes poses with cross-entropy optimization, while RoboSnap~\citep{zhang2026robosnap} alternates signed-distance optimization with settling. SimuScene~\citep{lee2026simuscene} uses simulated displacement to trigger shape stretching or regeneration. In \method, an agent identifies misplaced objects and calls geometry- and physics-grounded tools to correct them. These tools search for pose adjustments and test edits in simulation, returning feedback to guide further inspection and correction.

\noindent\textbf{Simulation for Policy Evaluation.}
\citet{kadian2020sim2real} tune simulation to improve sim--real correlation. SIMPLER~\citep{li2024simpler} calibrates controllers from trajectories and matches backgrounds and textures. PolaRiS~\citep{jain2025polaris} combines Gaussian splatting with policy co-training on real and simulated data; REALM~\citep{sedlacek2026realm} calibrates robot dynamics through replay and tests visual, semantic, and behavioral generalization. Following these, we assess our system through real-sim policy rollout agreement, task success-rate correlation, and open-loop trajectory replay.

\section{Method}
\label{sec:method}

Given a single RGB image, we reconstruct an editable, simulation-ready 3D scene that matches its appearance and object arrangement. The system primarily targets tabletop scenes for robotics and can also process room scenes by treating the floor as the main support surface.

\begin{figure}[t]
    \centering
    \vspace{-1mm}
    \captionsetup{font=small}
    \vspace{-3mm}
    \includegraphics[width=\linewidth]{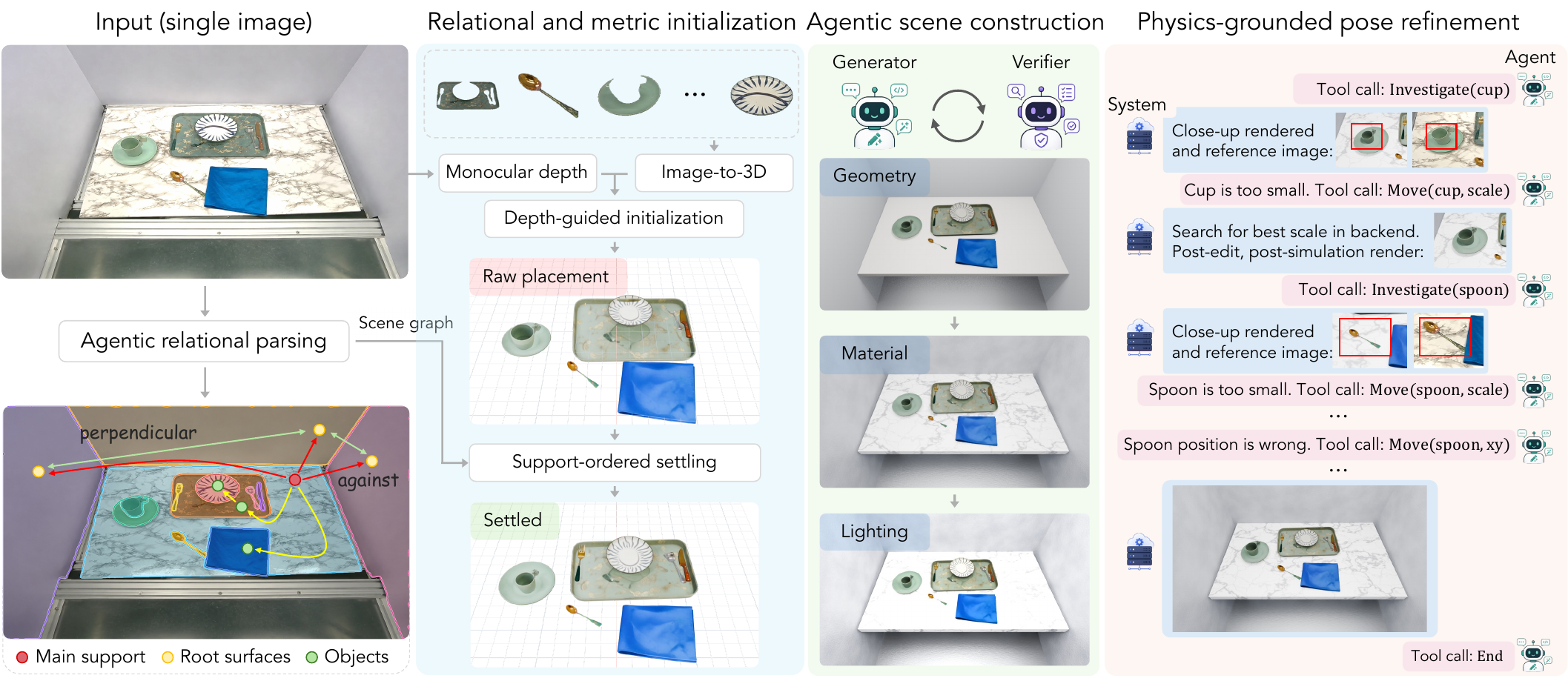}
    \vspace{-6mm}
    \caption{\textbf{Overview of \method.}
    From a single image, \method recovers objects, supporting surfaces, and their relations, and initializes a gravity-aligned metric scene through physical settling. Geometry, material, and lighting agents construct the surrounding scene. Finally, a pose-refinement agent inspects paired reference/render crops and invokes geometry- and physics-grounded tools to correct object placement.
    }
    \vspace{-3mm}
    \label{fig:pipeline}
\end{figure}

Let $\mathcal{O}$ denote movable objects and $\mathcal{R}$ denote root surfaces, such as tables, floors, and walls. Each entity has a persistent identity and a state defined by
\begin{equation}
\label{eq:entity-state}
x_v=\left(M_v,\tau_v,\theta_v^{\mathrm{phys}},X_v\right),
\qquad v\in V=\mathcal{O}\cup\mathcal{R}.
\end{equation}
Here, $M_v$ is geometry, $\tau_v$ appearance, $\theta_v^{\mathrm{phys}}$ collision and physical properties, and $X_v$ a metric transform with translation, rotation, and scale. Illumination is $L$, and the scene graph is
\begin{equation}
\label{eq:scene-graph}
G=(V,E),\qquad E=E_{\mathrm{sup}}\cup E_{\mathrm{surf}},
\end{equation}
where $E_{\mathrm{sup}}$ links objects to direct supports and $E_{\mathrm{surf}}$ records structural relations among roots.

Figure~\ref{fig:pipeline} summarizes the pipeline. We describe relational scene graph construction and scene object creation under Section~\ref{sec:preprocessing}, followed by Agentic scene construction (Section~\ref{sec:scene-refinement}) and physics-grounded object pose refinement (Section~\ref{sec:pose-refinement}).

\subsection{Relational and Metric Initialization}
\label{sec:preprocessing}

We first recover which entities are present, how they support one another, and where they lie in metric 3D. This relational and geometric evidence guides object reconstruction and physical settling.

\noindent\textbf{Relational parsing.}
A VLM proposes movable objects $\mathcal{O}$, root surfaces $\mathcal{R}$, and their support and structural relations, with a distinct description for each instance. MolmoPoint~\citep{molmopoint} localizes each entity from its description, and its predicted points select instance masks from SAM3~\citep{sam3} candidates for that category. Subsequent checks recover objects omitted from the initial inventory, repair split or merged masks, and revise root surfaces and their relations. For occluded objects, we optionally remove foreground occluders with LanPaint~\citep{zheng2025lanpaint} and re-segment the target. Entity identities persist throughout reconstruction (Appendix~\ref{app:relational-perception}).


\noindent\textbf{Metric initialization.}
The scene graph tells us what is present and how entities are related, but not their metric placement in 3D. We use MoGe-2~\citep{wang2025moge2} to estimate camera intrinsics and a dense metric 3D point map. When metric depth and calibrated intrinsics are available, such as from a robot's RGB-D camera, they replace these estimates. A plane fitted to the main support defines a gravity-aligned coordinate frame in which the support lies at \(z=0\). Fitted planes and estimated contact geometry also provide geometric evidence for validating the proposed scene relations. SAM3D~\citep{chen2026sam3d} reconstructs each movable object’s mesh and appearance from the image and its instance mask. We initialize its transform by translating and uniformly scaling the mesh to align with the corresponding metric 3D points. Finally, we estimate the physical properties for simulation following SceneSmith~\citep{pfaff2026scenesmith}.

\noindent\textbf{Support-ordered physical settling.}
With the support graph and metric object estimates in place, we settle the reconstructed objects while preserving their observed arrangement. Settling the full scene at once can cause objects to drift, topple, or resolve collisions in ways that differ from the image evidence. Thus, we settle objects in parent-before-child order under $G$. Each object is placed against its support and previously settled objects and simulated with PhysX in Isaac Sim. We then use ICP (adjusting translation, scale, and yaw) to improve alignment with its masked 3D observations and simulate it again.
Guarded fallbacks may adjust collision geometry and physical properties to obtain a resting configuration; Appendix~\ref{app:metric-initialization} details these adjustments and the validation procedure.

\subsection{Agentic Scene Construction}
\label{sec:scene-refinement}

With objects settled, we reconstruct root geometry $(M_r,X_r)$, materials $\tau_r$, and lighting $L$ in separate stages sharing a persistent Blender scene. Following SEIG~\citep{seig2026}, each stage edits one variable group. It pairs an editing agent with an independent verifier, both of which receive the scene graph, the current scene snapshot, and the permitted edits. The editor modifies scene code and renders the result, and the verifier compares the render with the reference image, then approves it or requests corrections within the same scope. Each revision continues from the updated scene, preserving earlier work across iterations and stages.

\noindent\textbf{Scene structure.}
The geometry agent constructs root surfaces in $\mathcal{R}$ as static colliders, positioning and orienting them with RANSAC-fitted planes and inferring their shapes and sizes from the input image; they must satisfy the required relations in $E_{\mathrm{surf}}$. Imported object geometry and appearance remain fixed; objects move only to correct measured contact defects or clear position mismatches.

VLM agents can produce visually plausible geometry that violates support or contact constraints, such as tables intersecting walls or objects floating above surfaces. Whereas VIGA and SEIG leave all checking to the verifier, we add a deterministic \texttt{check\_rules} gate that the editing agent must pass before verifier review.
The gate checks root placement, object support, and interpenetration, and reports the first failing level with the affected entities. It enforces a minimum geometric standard, leaving the verifier to assess visual alignment, including object placement and background geometry.


\noindent\textbf{Material and lighting.}
Once the structure is fixed, the material agent updates only root-surface materials and textures, including color, roughness, and surface patterns. The lighting agent then adjusts illumination and exposure to match the shadows, color, and contrast of the input image. Both stages leave background geometry, object placement, and object appearance unchanged, so pose refinement operates on fixed background geometry and rendering conditions.

\subsection{Physics-Grounded Object Pose Refinement}
\label{sec:pose-refinement}


After reconstructing the surrounding scene, we refine the object transforms $X_o$ to correct placement errors that remain after initialization.
A naive approach asks the VLM agent to adjust coordinates directly, e.g., by translating an object $0.05$ units in scene code.
However, lacking intrinsic metric grounding, agents often hallucinate these magnitudes, causing collisions or instability.
We therefore use an investigation and editing loop in which the agent diagnoses placement errors and calls tools to search for corrections and test them in simulation.

\noindent\textbf{Investigate--edit loop.}
A single agent session runs this loop in the persistent scene. Every object in $\mathcal{O}$ must be investigated at least once, and the agent may revisit objects as their surroundings change. The \texttt{Investigate} tool returns paired crops from the input image and current render, geometric measurements, and support information. Using this information, the agent decides which object to adjust and whether to change its position, scale, or orientation. For example, if an object looks too small in the render, the agent checks whether to move it closer to the camera or enlarge its mesh.

The agent passes the selected object and edit type to the \texttt{Move} tool, which supports horizontal translation, uniform scaling, and yaw adjustment. The tool determines the adjustment magnitude through bounded coarse-to-fine search, using the original pose as a baseline. Translation and scale scores combine silhouette overlap with depth; yaw combines centered silhouette overlap with DINOv2 features~\citep{oquab2024dinov2} from object crops. Appendix~\ref{app:pose-refinement} formalizes these scores in Eq.~\eqref{eq:pose-alignment-score}.

\noindent\textbf{Physical validation.}
For routine bounded moves, the \texttt{Move} tool then simulates the selected candidate only if it improves alignment. Physical checks reject capsize or excessive drift, and substantial new or worsened penetration. An update is retained only if these checks pass and its post-settling acceptance score matches or exceeds the baseline (Eq.~\eqref{eq:pose-update}); otherwise, the previous state is restored. If a support moves or is resized, its dependents in $E_{\mathrm{sup}}$ are repositioned and resettled as needed.

When bounded moves are insufficient, the agent can specify pose changes directly through the scene-code editing tool \texttt{Edit}. The tool settles affected objects in support order and returns the simulated render and physical diagnostics. The agent then decides whether to retain or undo the result. Appendix~\ref{app:pose-refinement} details search objectives and special cases.

\noindent\textbf{Context management and prompt caching.}
We summarize older interaction rounds in blocks, retaining a log of actions, scores, and outcomes and replacing outdated images with references to saved artifacts.
We retain the target image, stage-entry snapshot, and recent evidence. A per-object state table sits after the cache breakpoint, so measurements, pending checks, and budgets can change without invalidating the reusable prompt prefix. In our evaluation benchmark (Sec.~\ref{sec:experiments}), 84.1\% of pose-refinement input tokens are served from cache (Appendix~\ref{app:pose-refinement}; Figure~\ref{fig:app-prompt-caching}).

\noindent\textbf{Scene delivery and USD export.}
After refinement, a joint gravity simulation sets the final object transforms. We bake procedural materials and export meshes, textures, lights, and camera state to USD, preserving object identities, collision geometry, and physical properties for simulation. Appendix~\ref{app:settings-export} details conversion and simulator-side validation.

\section{Experiments}
\label{sec:experiments}

\begin{figure}[t]
    \centering
    \captionsetup{font=small}
    \vspace{-4mm}
    \includegraphics[width=\linewidth]{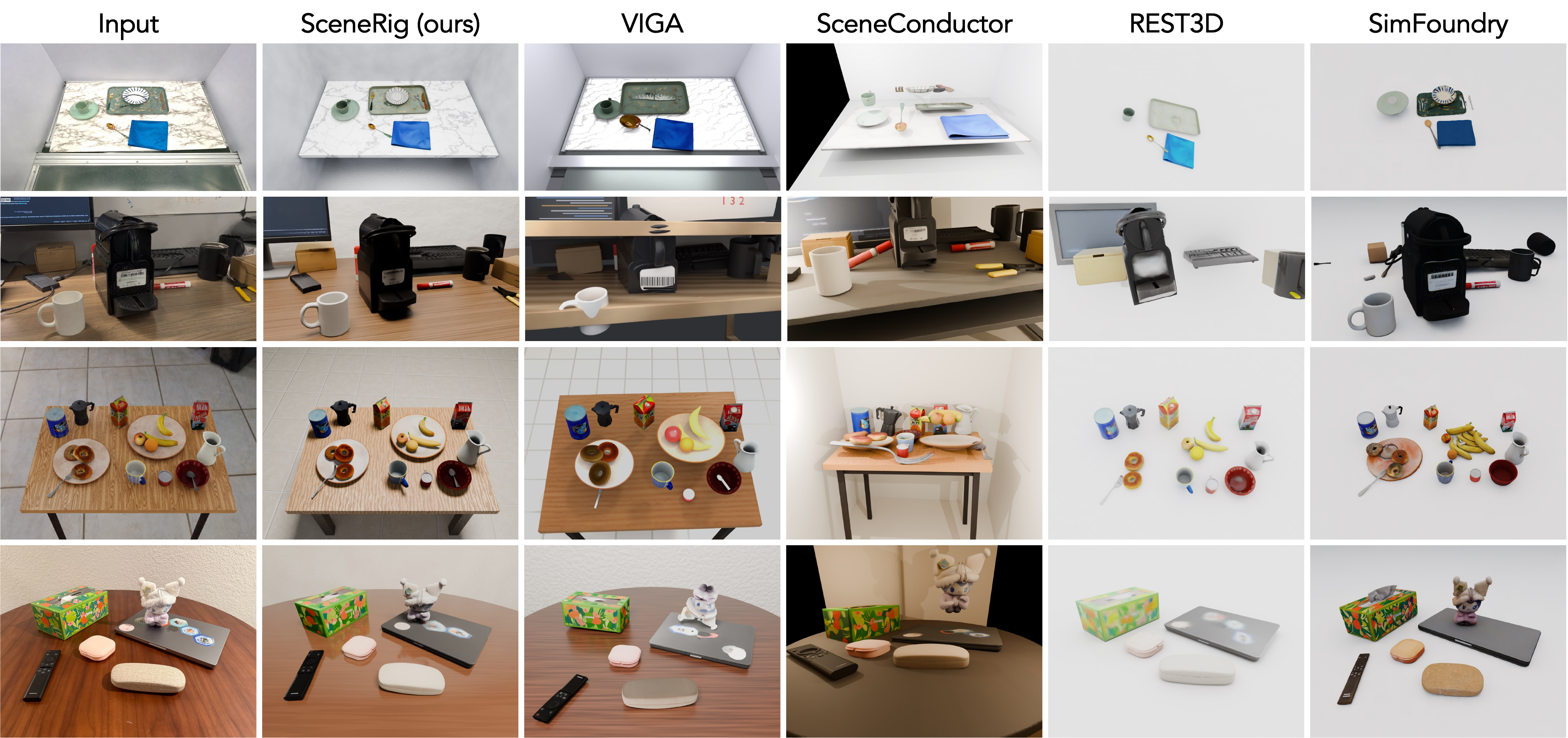}
    \vspace{-5.5mm}
    \caption{\textbf{Qualitative scene reconstruction results.}
    Compared with the baselines,
    \method more consistently preserves object identity, scale, and relative placement while reconstructing the surrounding scene.}
    \vspace{-3mm}
    \label{fig:qualitative-results}
\end{figure}

We evaluate \method's reconstruction fidelity, object recovery, 3D layout, and physical stability (Sec.~\ref{sec:reconstruction-quality}). We then examine physics-grounded placement and the effects of the VLM backbone and reconstruction harness (Sec.~\ref{sec:ablations}). Section~\ref{sec:robotics-applications} evaluates robotics applications; Appendix~\ref{app:additional-exp} provides additional analyses and results.

\subsection{Experimental Setup}

\noindent\textbf{Baselines.}
We compare with four baselines: VIGA~\citep{viga2026}, REST3D~\citep{ma2026rest3d}, SceneConductor~\citep{kim2026sceneconductor}, and SimFoundry~\citep{simfoundry2026}. To support object-centric evaluation and simulation, we augment VIGA's system prompt with a simple object-grouping contract: all parts of each movable object share a single empty Blender parent. We also evaluate SAM3D$^\ast$, which shares \method's segmentation, camera, objects, and scene context but retains SAM3D's predicted poses without our metric grounding, physical settling, or pose refinement. We use Claude Opus 5 as the default VLM wherever applicable. With GPT-6 Astra, we also compare against Codex, a general-purpose harness that reconstructs the scene directly in Blender without external models. Appendices~\ref{app:baseline-implementation}~and~\ref{app:additional-quantitative} provide implementation details and full harness results.

\noindent\textbf{Data.} Our benchmark contains 80 tabletop images, including 44 simulator renderings from RoboDojo~\citep{chen2026robodojo} and RoboLab~\citep{yang2026robolab} with ground-truth object identities and 3D positions. The remainder comprise real robot workspace images, Internet or AI-generated images, and smartphone photos. We evaluate source-view fidelity, perceptual preference, and physical stability on all scenes, and object-centric reconstruction and 3D layout on the simulator subset.

\noindent\textbf{Metrics.}
We report full-frame PSNR, SSIM, LPIPS, and nCLIP. Object-centric PSNR$^\ast$/SSIM$^\ast$ use object regions only (exclude background and table), while LPIPS$^\ast$/nCLIP$^\ast$ average the per-object crop scores. Instance metrics are adjusted Rand index (ARI), foreground IoU, and object-count MAE. For 3D layout, optimal subpattern assignment (OSPA)~\citep{schuhmacher2008ospa} at 5/10\,cm cutoffs penalizes position errors and unmatched objects; F1@10 uses one-to-one 3D matches within 10\,cm.  To account for differences in scale conventions across methods, we align each reconstructed layout to ground truth using a single global scale factor before computing OSPA and F1. GPT-5.6-terra and Qwen3.5-122B-A10B compare reconstruction pairs for completeness, layout, and appearance using the input image and both candidates' source- and nearby-view renders. We summarize preferences with mean-centered Bradley--Terry scores (inter-judge Cohen's $\kappa=0.87$). An object passes the stability test if its displacement after 5 s is below 50 mm and its final velocity is zero; a scene passes only if every object passes. Check details in Appendices~\ref{app:evaluation-metrics} and~\ref{app:vlm-physics-protocol}.

\subsection{Reconstruction Quality}
\label{sec:reconstruction-quality}

\begin{table}[t]
    \centering
    \small
    \vspace{-2mm}
    \captionsetup{font=small}
    \caption{\textbf{Scene-level reconstruction, pairwise preference, and physical stability.}
    VLM columns report mean-centered Bradley--Terry (BT) scores from pairwise judgments of source and nearby views. Within each criterion, a score advantage of $\Delta$ implies a modeled pairwise win probability of $1/(1+e^{-\Delta})$.
    An object passes the stability test if its displacement after 5\,s is below 50\,mm and its final velocity is zero; a scene passes only if every object does. All methods use Claude Opus 5 as the underlying VLM.}
    \vspace{-2mm}
    \resizebox{\textwidth}{!}{%
    \begin{tabular}{l|cccc|ccc|cc}
        \toprule
        & \multicolumn{4}{c}{Source-view reconstruction}
        & \multicolumn{3}{c}{Pairwise VLM evaluation}
        & \multicolumn{2}{c}{Physical stability} \\
        \cmidrule(lr){2-5}
        \cmidrule(lr){6-8}
        \cmidrule(lr){9-10}
Method & PSNR $\uparrow$ & SSIM $\uparrow$ & LPIPS $\downarrow$ &
nCLIP $\downarrow$ & Completeness $\uparrow$ & Layout $\uparrow$ &
Appearance $\uparrow$ & Scene (\%) $\uparrow$ & Object (\%) $\uparrow$ \\
        \midrule
VIGA & 11.414 & 0.529 & \underline{0.653} & 0.180 &
0.225 & \underline{-0.237} & \underline{-0.249} &
10.0 & 27.1 \\
REST3D & 11.092 & \textbf{0.590} & 0.729 & 0.214 &
-1.074 & -0.840 & -1.625 &
31.3 & 76.6 \\
SceneConductor & 10.140 & 0.503 & 0.727 & \underline{0.172} &
\underline{0.385} & -0.837 & -0.473 &
8.8 & 36.5 \\
SimFoundry & \underline{12.060} & \underline{0.582} & 0.720 & 0.194 &
-0.432 & -0.683 & -0.792 &
\underline{98.8} & \underline{99.9} \\
\textbf{\method} (ours) & \textbf{13.277} & 0.544 &
\textbf{0.518} & \textbf{0.122} &
\textbf{0.895} & \textbf{2.597} & \textbf{3.139} &
\textbf{100.0} & \textbf{100.0} \\
        \bottomrule
    \end{tabular}}
    \label{tab:source-view}
    \label{tab:vlm}
    \vspace{-2mm}
\end{table}

\begin{table}[t]
    \centering
    \captionsetup{font=small}
    \caption{\textbf{Object recovery and 3D layout.}
    Results on 44 simulator-grounded scenes. PSNR$^\ast$/SSIM$^\ast$ use object pixels (table excluded);
    LPIPS$^\ast$/nCLIP$^\ast$ average per-object crop scores with non-object pixels white.
    Layout metrics use global scale alignment; OSPA distances and cutoffs are in centimeters.}
    \vspace{-2mm}
    \label{tab:object-centric}
    \label{tab:layout3d}
    \resizebox{\textwidth}{!}{%
    \begin{tabular}{l|cccccccccc}
        \toprule
        & \multicolumn{7}{c}{Object-centric metrics}
        & \multicolumn{3}{c}{3D object layout} \\
        \cmidrule(lr){2-8}
        \cmidrule(lr){9-11}
Method & ARI $\uparrow$ & FG-IoU $\uparrow$ & Count MAE $\downarrow$ &
PSNR$^\ast$ $\uparrow$ & SSIM$^\ast$ $\uparrow$ &
LPIPS$^\ast$ $\downarrow$ & nCLIP$^\ast$ $\downarrow$ &
OSPA@5 $\downarrow$ & OSPA@10 $\downarrow$ & F1@10 $\uparrow$ \\
        \midrule
SAM3D$^\ast$ & \underline{0.784} & \underline{0.705} & -- &
\underline{12.810} & \underline{0.403} &
\underline{0.195} & \underline{0.078} &
\underline{2.7} & \underline{3.7} & \underline{0.88} \\
VIGA & 0.283 & 0.260 & 3.36 &
8.987 & 0.315 & 0.325 & 0.149 &
4.9 & 8.9 & 0.29 \\
REST3D & 0.210 & 0.204 & 3.20 &
9.636 & 0.378 & 0.296 & 0.148 &
4.7 & 8.7 & 0.29 \\
SceneConductor & 0.197 & 0.187 & 1.14 &
9.493 & 0.329 & 0.317 & 0.152 &
5.0 & 9.5 & 0.14 \\
SimFoundry & 0.676 & 0.589 & \underline{1.05} &
11.497 & 0.379 & 0.240 & 0.103 &
3.3 & 4.8 & 0.82 \\
\method (ours) & \textbf{0.878} & \textbf{0.819} & \textbf{0.27} &
\textbf{13.609} & \textbf{0.440} & \textbf{0.170} & \textbf{0.062} &
\textbf{2.4} & \textbf{3.1} & \textbf{0.93} \\
        \bottomrule
    \end{tabular}}
    \vspace{-4mm}
\end{table}

\noindent\textbf{Reconstruction fidelity and stability.}
\method achieves the best source-view PSNR, LPIPS, and nCLIP and the highest VLM preference scores across all three criteria (Table~\ref{tab:source-view}). VIGA and SceneConductor often fail the stability test, while REST3D's higher SSIM does not translate into accurate object recovery or placement. SimFoundry achieves high stability but trails in object appearance and 3D layout. \method leads all reported instance-recovery, object-appearance, and 3D-layout metrics (Table~\ref{tab:object-centric}), with every reconstructed scene passing the stability test.

\noindent\textbf{Qualitative results.}
Figure~\ref{fig:qualitative-results} compares reconstructions across diverse scenes. For visualization, REST3D and SimFoundry are shown with white support surfaces. In these examples, VIGA alters object geometry, while REST3D and SceneConductor misestimate object scale or placement. SimFoundry also introduces objects absent from the input. \method more faithfully preserves individual objects and their relative positions, including small utensils in cluttered scenes, while reconstructing the surrounding scene. Additional examples are provided in Appendix~\ref{app:additional-qualitative}.

\subsection{Ablation Studies}
\label{sec:ablations}

In this section, we evaluate our placement pipeline and the value of the bounded move tool within a physics-enabled agent and compare the influence of VLM backbones and different harnesses.

\begin{table}[t]
    \centering
    \begin{minipage}[t]{0.49\textwidth}
    \centering
    \footnotesize
    \captionsetup{font=small}
    \caption{\textbf{Physics-grounded object placement.}
    All components except placement and physical checks are shared. Post-hoc PhysX settles raw SAM3D poses for 5s; with code edits only, the agent sets move sizes.}
    \label{tab:placement-ablation}
    \vspace{-2mm}
    \setlength{\tabcolsep}{3pt}
    \resizebox{\linewidth}{!}{%
    \begin{tabular}{l|cc|cc}
        \toprule
        & \multicolumn{2}{c}{Instance structure}
        & \multicolumn{2}{c}{3D object layout} \\
        \cmidrule(lr){2-3}
        \cmidrule(lr){4-5}
Placement & ARI $\uparrow$ & FG-IoU $\uparrow$ &
OSPA@5 $\downarrow$ & F1@10 $\uparrow$ \\
        \midrule
SAM3D raw poses
    & 0.784 & 0.705 & 2.7 & 0.88 \\
\quad + post-hoc PhysX
    & 0.748 & 0.664 & 3.0 & 0.84 \\
\method, code edits only
    & 0.858 & 0.792
    & \textbf{2.4} & 0.92 \\
\method
    & \textbf{0.878} & \textbf{0.819}
    & \textbf{2.4} & \textbf{0.93} \\
        \bottomrule
    \end{tabular}}
    \end{minipage}\hfill
    \begin{minipage}[t]{0.49\textwidth}
    \centering
    \footnotesize
    \captionsetup{font=small}
    \caption{\textbf{Reconstruction harness comparison.}
    All methods use GPT-6 Astra. VIGA and Codex are evaluated after 5\,s of PhysX settling. Stability counts scenes that pass the stability test during this settle.}
    \label{tab:astra-harness}
    \vspace{-2mm}
    \setlength{\tabcolsep}{1.67pt}
    \resizebox{\linewidth}{!}{%
    \begin{tabular}{l|cc|cc|c}
        \toprule
        & \multicolumn{2}{c}{Object appearance}
        & \multicolumn{2}{c}{3D object layout} & Stability \\
        \cmidrule(lr){2-3}
        \cmidrule(lr){4-5}
        \cmidrule(lr){6-6}
Method & LPIPS$^\ast$ $\downarrow$ & nCLIP$^\ast$ $\downarrow$ &
OSPA@5 $\downarrow$ & F1@10 $\uparrow$ & Scenes $\uparrow$ \\
        \midrule
VIGA & 0.187 & 0.082 & 4.6 & 0.37 & 24/80 \\
Codex & 0.178 & 0.073 & 4.2 & 0.60 & 16/80 \\
\method & \textbf{0.151} & \textbf{0.060} & \textbf{2.3} & \textbf{0.95} & \textbf{80/80} \\
        \bottomrule
    \end{tabular}}
    \end{minipage}
    \vspace{-1mm}
\end{table}

\noindent\textbf{Physics-grounded object placement.}
All variants in Table~\ref{tab:placement-ablation} share the same components except object placement. The SAM3D variants retain raw poses, bypassing our metric initialization, settling, and pose refinement. These poses often penetrate or float above supports; settling them for 5\,s restores stability but reduces segmentation overlap and 3D placement accuracy. \method's support-ordered settling and pose refinement improve both while preserving stability. The code-edits-only variant removes \texttt{Move} while retaining \texttt{Edit}, which simulates agent-specified pose changes and returns feedback for agent review. This variant obtains worse object-appearance and 3D-layout scores (Tables~\ref{tab:placement-ablation} and~\ref{tab:app-ablation-placement}). Removing \texttt{Move} reduces robot trajectory replay success from 80\% to 44\% (Appendix~\ref{app:trajectory-replay-protocol}, Table~\ref{tab:replay-placement-ablation}). These results highlight the effectiveness of our physics-grounded tool design.

\noindent\textbf{VLM backbones and reconstruction harnesses.}
Within \method, replacing Opus 5 with GPT-6 Astra substantially improves appearance but only modestly improves 3D layout (Appendix~\ref{app:ablation}). With a shared GPT-6 Astra backbone, \method still achieves better results on most object-centric metrics, more accurate placement, and more stable scenes than VIGA and Codex (Tables~\ref{tab:astra-harness}~and~\ref{tab:app-astra-harness}). These results show that a stronger VLM alone does not ensure accurate and stable object placement: even when all methods use the same stronger backbone, \method still achieves more accurate placement and better physical stability.


\section{Robotics Applications}
\label{sec:robotics-applications}

\begin{table}[t!]
    \centering
    \begin{minipage}[t]{0.56\textwidth}
    \centering
    \captionsetup{font=small}
    \captionof{figure}{\textbf{Real-to-sim policy evaluation.}
    Each point is one policy--task pair (ten episodes); nested markers denote coincident pairs, and $r$ pools all ten pairs.}
    \label{fig:policy-correlation}
    \vspace{-2.5mm}
    \includegraphics[width=\linewidth]{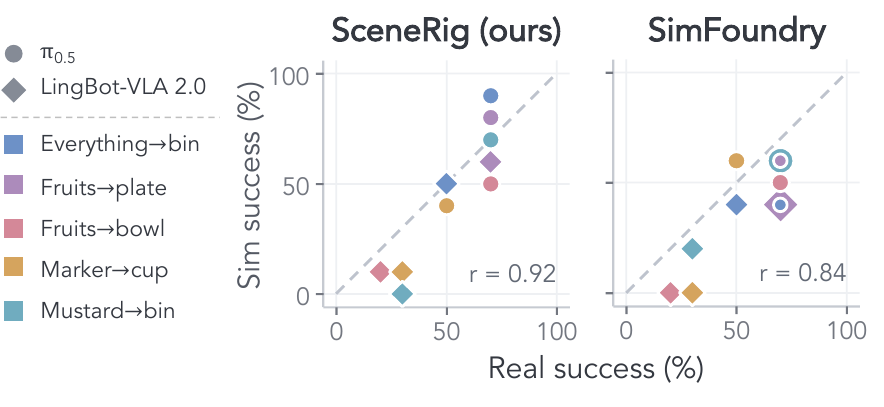}
    \end{minipage}\hfill
    \begin{minipage}[t]{0.42\textwidth}
    \centering
    \small
    \setlength{\tabcolsep}{2pt}
    \renewcommand{\arraystretch}{1.1}
    \captionsetup{font=small,justification=raggedright,singlelinecheck=false}
    \caption{\textbf{Open-loop trajectory replay.} Five successful real demonstrations per task, with shared commands and simulation settings.}
    \label{tab:replay}
    \vspace{-2mm}
    \begin{tabular}{lcc}
        \toprule
        Task & \method & SimFoundry \\
        \midrule
        Everything$\to$bin & \textbf{3/5} & 0/5 \\
        Cup$\to$bowl & \textbf{4/5} & 3/5 \\
        Fruits$\to$bowl & \textbf{4/5} & 2/5 \\
        Marker$\to$cup & \textbf{4/5} & 0/5 \\
        Mustard$\to$bin & \textbf{5/5} & 4/5 \\
        \midrule
        All & \textbf{20/25} & 9/25 \\
        \bottomrule
    \end{tabular}
    \end{minipage}
    \vspace{-4mm}
\end{table}

We assess \method with real-to-sim policy evaluation and open-loop trajectory replay. We compare with SimFoundry, which reconstructs the source-view camera and accounts for physical stability; REST3D's exported scenes do not retain a calibrated source camera, while VIGA and SceneConductor omit physical simulation. For each episode, both methods reconstruct the scene from an armless head-camera image using its metric depth and calibrated intrinsics instead of monocular estimates. Scenes are registered to the robot using the calibrated head-camera pose.


\subsection{Real-to-Sim Policy Evaluation}
\label{sec:sim2real-eval}

\textbf{Setup.} We evaluate finetuned $\pi_{0.5}$~\citep{pi05} and LingBot-VLA 2.0~\citep{lingbotvla2} on five tabletop tasks, using separate sets of 50 real episodes per policy (10 episodes per task, 100 episodes in total). Both reconstructions of each episode run the policy checkpoint used on the real robot, sharing simulator, calibration, and global simulation settings. We report success rates, episode-level agreement, and Cohen's $\kappa$ for each policy. Pearson's $r$ is computed both within each policy across five tasks and across all ten policy–task pairs. Details are provided in Appendix~\ref{app:sim2real-protocol}.

\textbf{\method better predicts real-robot outcomes.} Simulated outcomes match real ones in 69\% of episodes with \method versus 54\% with SimFoundry (Tables~\ref{tab:sim2real}~and~\ref{tab:lingbot-sim2real}), a significant paired difference (McNemar $p=0.017$), and \method's Cohen's $\kappa$ exceeds SimFoundry's by 0.29 (95\% CI [0.08, 0.50]). Across ten policy--task pairs, real and simulated success rates correlate at $r=0.92$ with \method and $0.84$ with SimFoundry, with mean maximum rank violations (MMRV)~\citep{li2024simpler} of 0.04 and 0.11 (Figure~\ref{fig:policy-correlation}; Table~\ref{tab:robotics-stats}).

\subsection{Open-Loop Trajectory Replay}
\label{sec:trajectory-replay}

\noindent\textbf{Setup.}
We collect and evaluate 25 successful teleoperated demonstrations, five per task. Both \method's and SimFoundry's reconstructions of each episode are imported into Isaac Sim with the calibrated bimanual robot. Controllers track the measured joint trajectory and gripper commands from the recorded initial configuration, without policy queries or trajectory adaptation. Commands, calibration, gains, and physics settings are shared (Appendix~\ref{app:trajectory-replay-protocol}).

\begin{figure}[t!]
    \centering
    \captionsetup{font=small}
    \IfDefinedSwitch{\enableReplayAnimation}{%
        \animategraphics[
            width=\linewidth,
            autoplay=false,
            autoresume=false,
            autopause=false,
            loop=false,
            controls={play,step,stop},
            controlsaligned=center,
            buttonsize=1.4em,
            poster=first,
            timeline={figures/videos/replay-5256fc52/timeline.txt},
            type=jpg,
            alttext={SceneRig and SimFoundry reconstruct the same input image. Each pair shows the reconstruction at its initial state and the replay result at its final state. SimFoundry fuses the apple and plate, so grasping the apple lifts the plate. Click Play for synchronized real and simulated execution.}
        ]{8}{figures/videos/replay-5256fc52/replay-}{0000}{0099}%
    }{%
        \includegraphics[width=\linewidth]{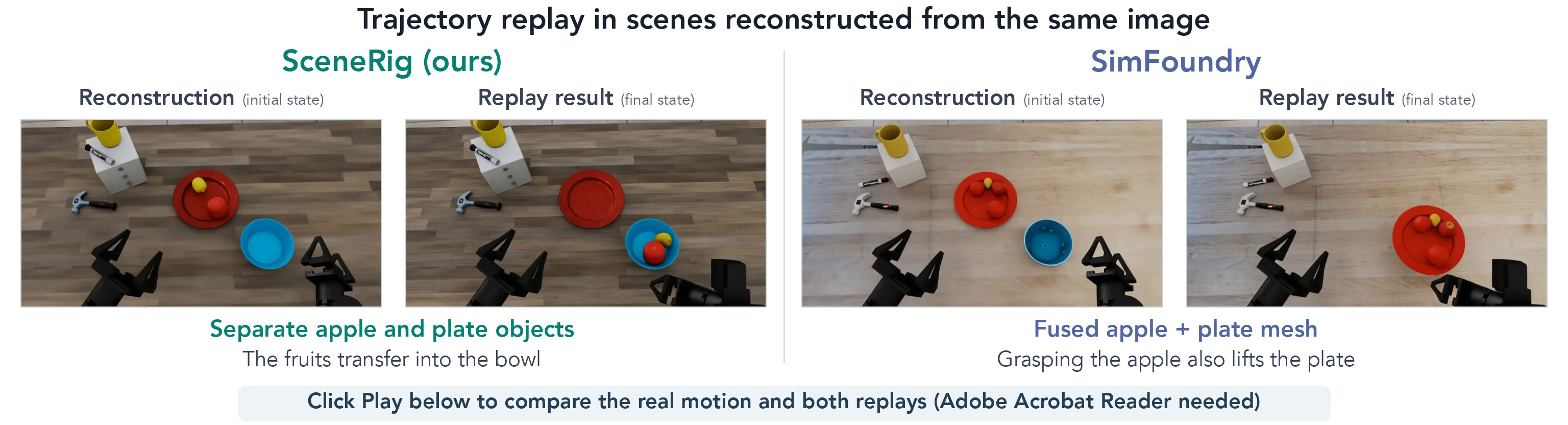}%
    }
    \vspace{-3mm}
    \caption{\textbf{Open-loop trajectory replay in reconstructed scenes.} Both methods reconstruct the same image. Each pair shows the initial reconstruction and the final result of replaying identical recorded joint and gripper commands. \method transfers the fruits separately; SimFoundry's fused apple--plate mesh causes the plate to be lifted with the apple. View the embedded animation in \textit{Adobe Acrobat Reader} or visit our~\page.}
    \vspace{-3mm}
    \label{fig:trajectory-replay}
\end{figure}

\noindent\textbf{\method more reliably supports open-loop trajectory replay.}
\method successfully replays 20/25 demonstrations, versus 9/25 for SimFoundry, with higher success on every task (Table~\ref{tab:replay}); it succeeds on every demonstration that SimFoundry replays successfully (exact McNemar $p=0.001$). SimFoundry fails due to missing objects, misplaced containers, and geometry that obstructs the robot. In Figure~\ref{fig:trajectory-replay}, its fused apple--plate mesh lifts the plate when the apple is grasped, while \method preserves separate objects and supports the recorded manipulation. 

\begin{table}[t!]
    \centering
    \footnotesize
    \setlength{\tabcolsep}{3pt}
    \captionsetup{font=small}
    \caption{\textbf{Real-to-sim policy evaluation.} Results for $\pi_{0.5}$ on 50 real episodes and their reconstructions. Entries count successful episodes (success) or matching real and simulated outcomes (agreement). Cohen's $\kappa$ uses these 50 episodes; Pearson's $r$ correlates the five real and simulated task success rates.}
    \vspace{-2mm}
    \resizebox{\textwidth}{!}{%
    \begin{tabular}{p{2cm} l c M{2.0cm} M{2.0cm} M{2.0cm} M{2.0cm} M{2.0cm}|cc}
        \toprule
        Environment & & Everything$\to$bin  & Fruits$\to$plate & Fruits$\to$bowl & Marker$\to$cup & Mustard$\to$bin & All & Cohen's $\kappa\uparrow$ & Pearson $r\uparrow$ \\
        \midrule
        Real-world & \multicolumn{1}{|l|}{success}   & 7/10 & 7/10 & 7/10 & 5/10 & 7/10 & 33/50 & -- & -- \\
        \midrule
        \multirow{2}{*}{SimFoundry} & \multicolumn{1}{|l|}{success}   & 4/10 & 6/10 & 5/10 & 6/10 & 6/10 & 27/50 & -- & -0.38 \\
                   & \multicolumn{1}{|l|}{agreement} & 5/10 & \textbf{7/10} & 2/10 & 5/10 & 5/10 & 24/50 & -0.07 & -- \\
        \midrule
        \multirow{2}{*}{\textbf{\method}}   & \multicolumn{1}{|l|}{success}   & 9/10 & 8/10 & 5/10 & 4/10 & 7/10 & 33/50 & -- & \textbf{0.70} \\
                   & \multicolumn{1}{|l|}{agreement} & \textbf{8/10} & \textbf{7/10} & \textbf{6/10} & \textbf{7/10} & \textbf{8/10} & \textbf{36/50} & \textbf{0.38} & -- \\
        \bottomrule
    \end{tabular}
    }
    \vspace{-1mm}
    \label{tab:sim2real}
\end{table}

\section{Discussion}
\label{sec:discussion}


\textbf{Conclusions.} We present \method, an agentic system that reconstructs simulation-ready scenes from single images. Its modular harness equips agents with geometric and physical tools to refine object placement through edit-and-validate loops. Across diverse tabletop scenes, \method improves visual fidelity and placement accuracy while maintaining physical stability. Its scenes better predict real-world policy outcomes and more reliably reproduce recorded manipulations. These results offer a path to scaling robot policy evaluation by bringing real workspaces into simulation.


\textbf{Limitations and future work.} Although \method accurately places objects in tabletop scenes, complex room scenes remain challenging due to occlusion, depth ambiguity, and frequent contacts. SAM3D's lack of PBR materials limits photorealism, and our current representation treats each reconstructed object as a single rigid body and does not recover joint structures or deformation. Future work will explore improved depth estimation and image-to-3D models.

\arxivSwitch{}{\subsection*{AI use statement}

We used generative AI tools to assist in implementing the authors' ideas,
running experiments, collecting and organizing results, and polishing the
manuscript for clarity and readability. AI-based image generation, scene
reconstruction, and evaluation used in this work are described in the method
and experimental sections and the appendix. The authors retain responsibility
for the correctness of the implementation and results and for all claims and
content in this paper.

\subsection*{Reproducibility statement}

We provide implementation and evaluation details to support reproducibility, although LLM outputs may vary across runs. Appendix~\ref{app:implementation-details} describes the full pipeline, including the models (Table~\ref{tab:app-models}) and implementation settings (Table~\ref{tab:app-settings}), and Appendix~\ref{app:experimental-details} specifies the baselines, metrics, and robotics protocols. \arxivSwitch{Our code is available on the~\page.}{We will release our code upon acceptance.}
}

\bibliographystyle{plainnat}
\bibliography{bib/main}
\clearpage
\appendix

\section{Supplementary Material Overview}
\label{app:supplementary-overview}

This supplementary material provides implementation details, evaluation protocols, and additional results for \method.
Section~\ref{app:implementation-details} describes relational perception, object initialization and support-ordered settling, agentic scene construction, object pose refinement, implementation settings, and simulator export.
Section~\ref{app:experimental-details} details baseline implementation and metric definitions, as well as the protocols for real-to-sim policy evaluation and open-loop trajectory replay.
Section~\ref{app:additional-exp} presents ablation studies on pose/physics modules, VLM backbones, and harness comparisons, including the direct Codex baseline with GPT-6 Astra, and additional qualitative examples.
\arxivSwitch{Code is available at \expandafter\url\expandafter{\arxivCodeURL}.}{Accompanying this file is our anonymous~\page.}

\section{Additional Implementation Details}
\label{app:implementation-details}

\begin{figure}[t]
    \centering
    \vspace{-3mm}
    \includegraphics[width=0.8\linewidth]{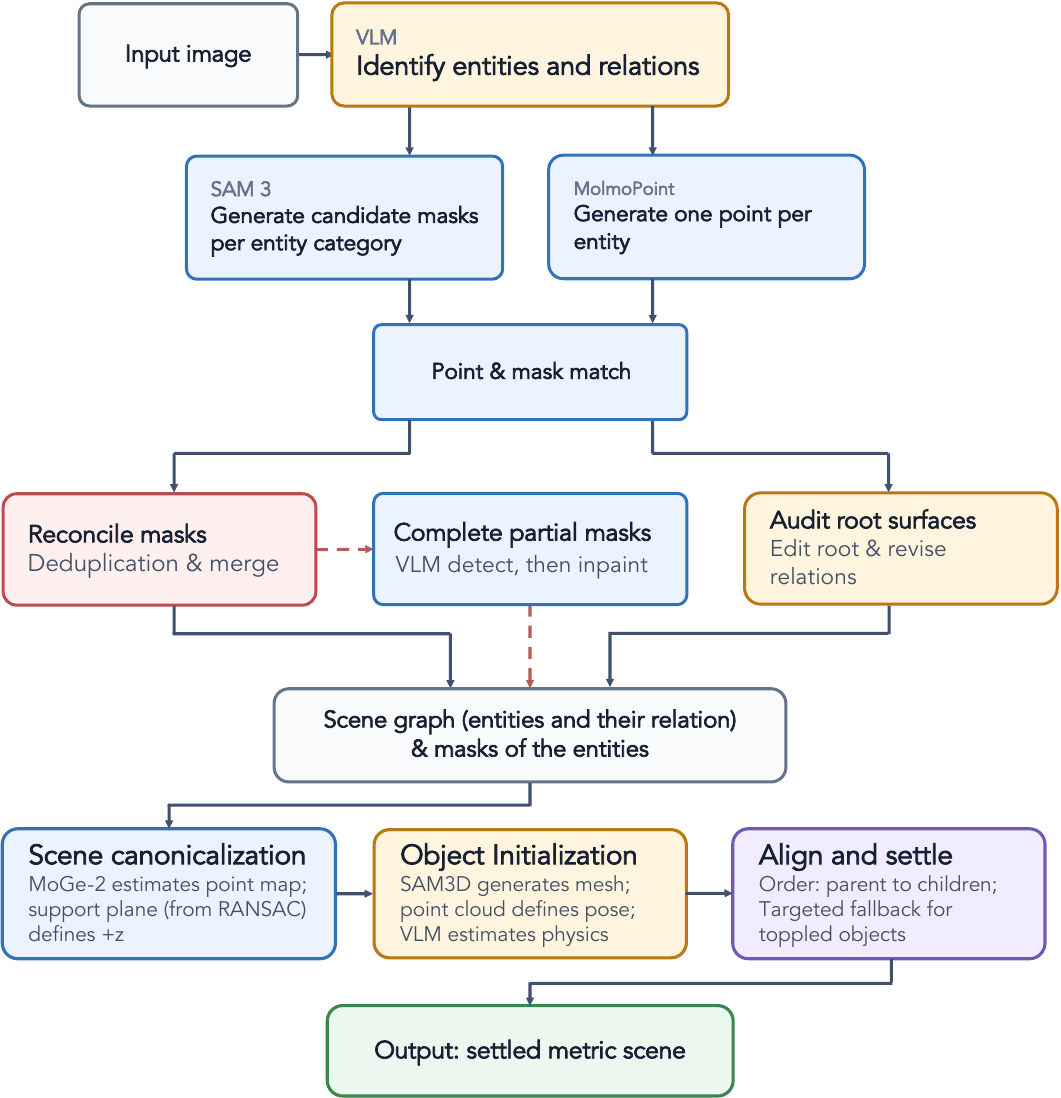}
    \caption{\textbf{Relational perception and mask recovery.} A VLM identifies
    entities and proposes relations.  MolmoPoint locations select SAM3 masks;
    subsequent checks recover missed objects, merge duplicate or split masks,
    complete occluded objects, and revise the root-surface records.}
    \label{fig:app-relational-perception}
\end{figure}

\begin{figure}[htbp]
    \centering
    \includegraphics[width=0.9\linewidth]{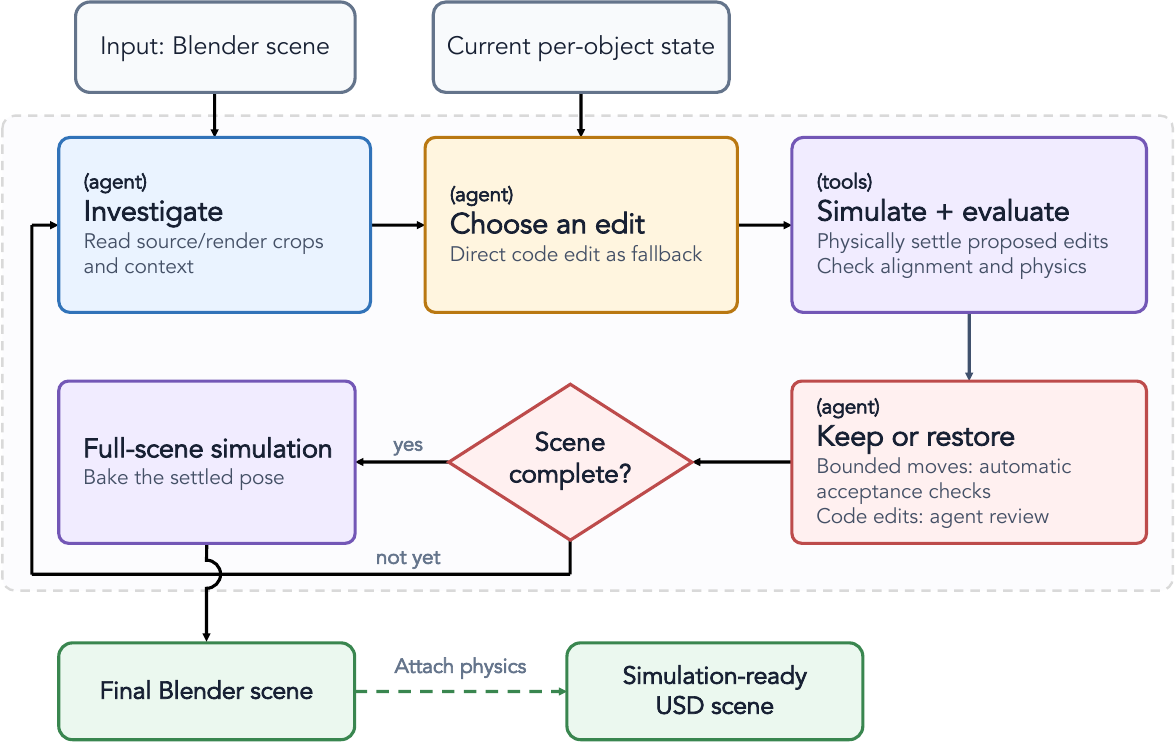}
    \caption{\textbf{Pose refinement and export.} The agent investigates
    objects and requests edits. The \texttt{Move} tool searches and physically validates bounded moves; the \texttt{Edit} tool simulates direct edits for agent review.
    After a final gravity pass, object poses and physical properties are exported to a simulation-ready USD file.}
    \label{fig:app-pose-refinement}
\end{figure}

This appendix expands the perception, initialization, and refinement procedures
in Sections~\ref{sec:preprocessing}--\ref{sec:pose-refinement}.
Figures~\ref{fig:app-relational-perception}, \ref{fig:app-stage-refinement},
and~\ref{fig:app-pose-refinement} summarize relational perception, agentic
scene construction, and the pose-refinement loop, respectively.
We use the entity IDs and graph notation defined in Section~\ref{sec:method}
throughout.  The descriptions below distinguish model predictions, deterministic
tests, and fallbacks.  Tables~\ref{tab:app-models} and~\ref{tab:app-settings}
collect the model roles and implementation defaults.

\subsection{Relational perception and mask recovery}
\label{app:relational-perception}

\noindent\textbf{Scene routing and entity parsing.}
A routing VLM first classifies the image as a room, a close-up showing a table, or a tabletop close-up. The effective route selects the proposer/verifier prompt pair for support construction and connectivity filtering. 

A VLM call then groups entities by category.  Each category
record contains a short detector-friendly noun, up to three synonyms, and a
list of instances.  For example, a single \texttt{mug} record can contain a
white mug at the front and a blue mug behind it.  Each instance has its own
description, type (movable object or root surface), and proposed direct
support.  Attributes and relative locations belong in the instance description,
not in the category name.  Category names are normalized and duplicate category
records are consolidated.  The parsing pipeline assigns category-local IDs such as \texttt{mug\#0} and \texttt{mug\#1}; later recovery steps preserve surviving IDs and allocate new ones for additional instances. Root-surface categories are
recognized by their head noun, such as table, wall, floor, or counter. The
proposer also emits relations among these surfaces, such as a wall behind a table.

The routing distinction between a tabletop close-up and a view of the table's
body or base determines whether the main support needs only a slab or connected
structure beneath it.  The ordinary close-up path focuses on the
main support and its objects, excluding people and objects on unrelated
surfaces.  An optional room-scale mode instead treats the floor as a root and
furniture as objects.  Containers remain objects, with their contents linked
to the container as their direct support.  Bounding walls receive separate
root IDs.  Following the proposed support chains identifies the root holding
the most objects as the main support.  Category grouping shares detection
work; it does not assume that all instances have equal physical size.

\noindent\textbf{Point-guided mask selection.}
MolmoPoint~\citep{molmopoint} localizes each instance from its description,
including root surfaces.  SAM3~\citep{sam3} produces a full-image mask pool for
each category; instances of that category share the pool but select masks
using their own points. Selection starts with the smallest mask containing
the point, then prefers a larger candidate if it contains at least $80\%$ of
that mask.
Partially overlapping instances remain separate. Synonym pools are computed as needed and cached. The recovery ladder proceeds from full-image text prompts, to text prompts on a point-centered crop, to point-prompted SAM3 with all of the instance's localized points as foreground prompts. Crops initially span one third of the image width and height. A mask cut by an internal crop boundary triggers a larger, two-thirds crop.  Root surfaces skip cropping because a local crop can lose their full extent.

Selection is instance-aware.  A candidate covering three or more distinct
object points is rejected as a merged-object mask; declared support children
are excluded from this test.  For a singleton category, compatible fragments
near the localization point can be joined, and a substantially larger synonym
mask can replace a detected part with the whole object.  Point prompting normally requires SAM3's predicted IoU to reach $0.6$.  If single-point prompting returns plausible masks with more than a twofold area difference, a VLM selects the intended object extent instead of comparing confidence scores across these scales; this scale-adjudicated choice bypasses the confidence floor. Roots have a more permissive final fallback: a nearby point-to-mask match is allowed, and a largest-mask choice is used only when localization returns no point.  Root point-prompt masks are not subject to the object confidence floor. Each accepted mask records its recovery tier. Instances that exhaust the ladder without a usable mask enter the dropped list with a reason.

\noindent\textbf{Inventory and overlap checks.}
Object-level checks target missed instances in repeated categories, such as
masking only two of three mugs, and objects split across masks, such as a
monitor separated into its base and screen.  These checks reconcile duplicate
assignments and overlaps.

Two instance descriptions can localize the same object.  When this causes a
duplicate or failed assignment, MolmoPoint gets one retry on an image with
already claimed instances of that category crossed out.  SAM3 still segments
the original image.  For categories initially containing multiple objects,
unused masks in the category/synonym pools trigger a plural localization query,
such as ``all the mugs.''  A new instance is accepted only when a point from
the plural query falls inside a distinct unclaimed SAM3 mask and the mask
passes the overlap checks against claimed objects; a VLM supplies its description.
Roots are excluded.  An initially singleton category is recounted only when
its mask combines multiple instances: distinct SAM3 members must each occupy
$8$--$92\%$ of its area, lie at least $80\%$ inside it, and together cover at
least $70\%$ of it.  MolmoPoint must localize at least two members
with a plural query.  One member retains the original ID; the others enter
the same unclaimed-mask acceptance procedure.

High-overlap masks are first deduplicated by entity type (IoU $>0.7$ or containment $>0.8$).  Cross-category containment is deferred
because it may represent a vessel and its contents. For touching object masks, the merge cascade first joins high-overlap duplicates at IoU $\geq0.45$, except deferred containment pairs, which require
VLM review. Below this threshold, same-category pairs are never merged.
Cross-category pairs with IoU $0.25$--$0.45$ are sent to a VLM; other touching pairs are skipped. Mere contact does not imply one object.
The VLM sees an unmarked crop and a red/blue mask overlay. Accepted merges
union the masks, preserve the survivor's ID, and redirect support references.
Containment is the fraction of the smaller mask covered by the larger.
When over $80\%$ of a smaller, cross-category object is embedded in a larger
mask, its pixels are removed from the larger mask unless this would remove over half of that mask. This carving runs before occlusion completion; legitimate overlap between completed masks is not carved away afterward.

Object components containing their localization points are retained.  A fragment of at least 50 pixels can be restored if it reconnects through an occluder's mask, contains no other object's point, and overlaps other objects by at most $10\%$ of its area.

\noindent\textbf{Root-surface verification.}
A separate VLM receives the original image, a color-coded root-mask overlay
labeled with IDs, and the root descriptions and proposed relations.  It
corrects the relations and describes missing surfaces.  It can add a root by description without rerunning segmentation, but cannot add, remove, or modify movable objects. Initially unsegmentable roots are dropped; this review can reintroduce missing surfaces as description-only roots in $\mathcal{R}$ for visual construction.

\noindent\textbf{Occlusion completion.}
Preprocessing invokes generative resegmentation after segmentation; its
per-instance checks determine whether an image edit or mask recovery is needed. Mask pairs with at least 20 overlapping pixels after three-pixel dilation are possible occluder--target pairs.  A VLM uses their image crops and depth-order evidence to identify the foreground object. A separate crop check for each instance estimates visibility and frame truncation and identifies missing structural parts, such as a handle or base. These judgments feed a deterministic recovery policy.  Self-occlusion is not a missing external part.  A supporting ancestor will not be treated as an occluder of its descendant. Completion is considered for a missing structural part, visibility below $75\%$, or an overlap-separation loss of at least 150 pixels or $3\%$ of the target mask.

We apply LanPaint~\citep{zheng2025lanpaint} with
Qwen-Image-Edit-2509~\citep{qwenimageedit2509} to remove the occluder and its support descendants.  The edit mask is their union, dilated by 12 pixels; the prompt names the objects to remove and the target to reveal.  SAM3 resegments the edited image using a part-aware description, the category, the original localization point, and finally synonyms.  Candidate masks must have IoU at least $0.25$ with the original target or reconnect a missing part within a 10-pixel dilation of it; adjacent-part unions cannot exceed eight times the original area.  Sibling masks are excluded.  If a candidate covers over $80\%$ of a removed foreign object's footprint, that footprint is clipped outside the original target mask.  Removed objects in the target's support subtree are exempt.  The chosen mask must grow by more than $2\%$. When overlap carving alone triggers recovery, failed resegmentation can instead use the union of the original target and removed footprints.  Other failures retain the original evidence.  If only a structural part was missed, resegmentation uses the original image without an edit.

A VLM audits newly generated candidates using aligned before/after RGB images
and masked cutouts.  It checks target identity and count and mask fidelity;
edited images additionally require preserved observed geometry and plausible
completion, with occluder removal checked when applicable.  All applicable
checks must pass; rejection, uncertainty, or an unavailable judgment retains
the original evidence.  This audit also applies to the union fallback and
border completion.  Older cached completions without an audit may be reused,
but are explicitly marked unverified.

RGB completion alone would leave the removed object's depth in the newly
revealed region.  We therefore invalidate depth in the edit hole and pass
the edited RGB image, remaining MoGe-2 depth, and camera intrinsics to
LingBot-Depth~\citep{tan2026lingbotdepth}.  Its completed metric points support
mesh reconstruction and placement.  Original masks, depth, and camera records
are retained separately, and failed image or depth completion falls back to
the available original evidence.

A separate border-completion branch handles masks reaching within three pixels
of exactly one left or right image edge.  Completion requires either a VLM
estimate of at least $10\%$ beyond-frame extent or edge contact in at least
$85\%$ of the rows within the mask's bounding box.
Eligible cases receive a $25\%$ canvas extension on that
side, followed by LanPaint/Qwen outpainting and LingBot depth completion.
Occluder removal, when needed, is a separate edit.  Padding shifts the camera's
principal point but does not change its physical viewpoint.  The extended
image helps reconstruct the complete asset; source-view registration still
uses the original image domain.  This branch does not extrapolate arbitrary
top, bottom, or two-sided truncations.

\noindent\textbf{Post-completion cleanup.}
After completion, a further VLM pass checks whether
nearby point-prompted detections are parts of one physical assembly, such as
a stand and its supported device.  Candidate masks must touch or lie within
a small dilation of one another; proximity alone does not justify merging
them.  The VLM must identify them as parts of the same object.  Merges update
support references as well as masks.  Objects without a usable mask are removed
before the final inventory is fixed, with their records and reasons retained
for inspection.

After completion and component merging, a connectivity check filters objects
assigned directly to masked roots.  An object must connect to its support,
either directly or through accepted objects sharing that support, within
approximately $3\%$ of the image's shorter dimension.  Disconnected objects
are removed before fixing the inventory.  The check uses completed masks
when available, except for border completion, whose extended canvas requires
the original in-frame mask.  Objects supported by other objects or unmasked
roots are exempt from this check.


\subsection{Metric initialization and support-ordered settling}
\label{app:metric-initialization}

\noindent\textbf{Metric frame and relation validation.}
MoGe-2~\citep{wang2025moge2} estimates metric depth, camera intrinsics, normals,
and a dense point map.  A RANSAC fit to the main support estimates the upward
direction.  The fit starts from the localized support component and includes
compatible coplanar fragments.  A direction-consensus step also considers
planar support components and planes on directly supported objects, reducing
the effect of a contaminated support mask.  The mean predicted support normal
is a fallback when a point-based fit is unreliable.  Small estimated camera
rolls below $2.5^\circ$ are removed in the monocular path; calibrated depth
inputs retain their measured roll.

For an upward unit vector $u$ and a support-plane point $p_0$, the canonical
transform satisfies
\begin{equation}
\label{eq:app-canonical-frame}
R_g u=e_z,\qquad p^{\mathrm{can}}=R_g p+t_g,\qquad
t_g=(0,0,-e_z^\top R_g p_0)^\top.
\end{equation}
Thus gravity is along $-z$ and the main support lies at $z=0$.
The same transform is applied to the camera.  If the declared main support
has no usable mask, another supported horizontal root is considered; if none
is measurable, a fallback level ground is placed below the observed object
bases.  These fallback choices are recorded rather than treated as measured
support planes.

Root-plane reliability depends on both RANSAC inliers and image coverage, so a narrow wall sliver cannot provide a strong orientation constraint.  Reliable near-horizontal normals are leveled and near-vertical walls are made plumb.  Ambiguous or unmeasurable planes remain visual construction tasks, not grounds for deleting a root.


Root relations distinguish flush support--wall contact (\texttt{against}),
a joined horizontal/vertical pair (\texttt{perpendicular}), two walls meeting
at a vertical edge (\texttt{corner}), and a lower surface supporting the full
base of an upper structure (\texttt{under}).
We then verify VLM-estimated relationships via monocular geometry. Confirmed relations become required construction constraints; uncertain or conflicting relations remain advisory, and rejected ones are not enforced.  The graph stores this status and its evidence along with $E_{\mathrm{sup}}$ and $E_{\mathrm{surf}}$.  Every support ID must resolve to an entity, parent and child records must agree, and support cycles are forbidden.

\noindent\textbf{Object assets and physical properties.}
SAM3D~\citep{chen2026sam3d} reconstructs a textured mesh from the
full-frame image and the instance mask. Its predicted orientation is transformed into the gravity-aligned frame without reflection. The mesh is centered at the metric position and uniformly scaled by the depth-projected 3D points of that object after outlier filtering. Both the posed mesh and an unposed canonical copy are saved; the latter supplies an alternative upright orientation during settling. Materials are rendered double-sided to avoid transparent-looking open mesh shells.

Following SceneSmith~\citep{pfaff2026scenesmith}, a VLM estimates material and mass from metric dimensions, sealed mesh volume, six mesh renders (four sides, top, and bottom), and a marked crop showing the object in the input image.  Friction comes from a material lookup table, not an unconstrained numerical VLM prediction.  An implied density above three times the material density triggers one reconsideration; a still-excessive estimate is clamped.  Low implied density is only flagged because hollow objects are legitimate.  Missing estimates use the documented density and friction defaults.  An estimate made at box dimensions $b_o^0$ is rescaled to
current dimensions $b_o$ by
\begin{equation}
\label{eq:app-mass-scale}
m_o^{\mathrm{phys}}=m_o^0\prod_{j=1}^{3}\frac{b_{o,j}}{b_{o,j}^0}.
\end{equation}
This gives the cubic law under uniform scaling. Physical-property overrides retain consistent
mass, center-of-mass, and inertia records.

\noindent\textbf{Nearby-view references.}
We combine SHARP~\citep{mescheder2026sharp} with
GPT-Image-2~\citep{openai2026gptimage2} to render nearby views.  SHARP predicts a 3D Gaussian
representation once from the input image, using the estimated focal length.
We render it at four source-camera offsets: $(+30^\circ,0)$,
$(-30^\circ,0)$, $(0,+15^\circ)$, and $(0,-15^\circ)$ in azimuth and elevation.
The orbit uses the estimated gravity direction and normally pivots about the
source viewing axis's intersection with the main-support plane. A grazing or
implausibly distant intersection falls back to the centroid of all objects' positions projected onto the support plane.

SHARP renders can contain holes, haze, or stretched edges. Thus, we use GPT-Image-2 to edit
each rendered image with instructions to repair these artifacts while preserving
clear content and viewpoint.  This is a mask-free edit of the rendered view,
not a text-only request for a new angle and not a paired input-photo edit.
We do not composite original pixels back into the generated result.  To detect
content drift, pixels with SHARP alpha below $0.8$, dilated by $1\%$ of the
image's long edge, are excluded from a comparison of raw and edited RGB.
The edit is rejected if the mean absolute difference over the remaining
observed pixels exceeds $25/255$, after resizing it to the raw render's
resolution. Up to three edit attempts are allowed.
An unsuccessful view may retain its raw render for diagnosis but is not exposed
as a trusted reference.

If SHARP fails, the fallback splats MoGe-2 points into the same cameras and
uses masked GPT-Image-2 editing to fill disocclusions.  Generated images are
visual guidance, not measured ground truth, and their hidden content may be
inaccurate. Agents receive the requested render/reference pair on demand; the
source-view reference is always the original photograph. The lighting stage
uses only that real source view, since generated illumination is not reliable
evidence for lighting estimation.

\noindent\textbf{Support-ordered settling.}
CoACD~\citep{wei2022coacd} converts each mesh into compound convex collision
shapes.  Supports and containers receive finer decomposition to preserve
interior cavities.  The validated decomposition is reused during pose
refinement and export when the mesh differs only by a recoverable transform;
geometry changes trigger rebuilding.  A persistent PhysX scene processes
objects in parent-before-child order under $E_{\mathrm{sup}}$, with larger
footprints prioritized where dependencies permit.  This order applies to
initial placement, not the pose agent's investigation sequence.  Temporary
root slabs provide support until the initialization agent builds
root geometry.

Each object is first dropped against its support context, then against the
already placed scene.  It is released 2\,mm above collision-free contact and
simulated with all six rigid-body degrees of freedom.  For a non-rollable
object, excessive tilt triggers a recovery ladder: try the canonical upright
mesh orientation, then a guarded physical-property adjustment.  The adjustment
can flatten a narrow collider base band, move or lower the center of mass,
and increase friction and damping, with matching inertia.  A sliver-like
contact footprint disables this adjustment to avoid manufacturing an implausible
balancing pose.  If no upright candidate works, the actual fallen resting pose
is retained and flagged.  Rollable objects instead use horizontal displacement
to detect instability; rolling in place is allowed, and center-of-mass
stabilization is skipped.  Small lateral searches address collisions with
neighbors, and a bounded inward nudge can move an object off its support's rim.

After settling, trimmed iterative closest point (ICP) alignment fits the mesh
to the masked metric points.  It optimizes horizontal translation, yaw, and
uniform scale about the object's bottom center; the fitted vertical offset is
discarded so simulation determines resting height.  The correction must reduce
point-to-mesh RMS error by at least $20\%$.  Scale is fixed for equal-size or
otherwise scale-locked assets.  ICP is skipped after a successful rim or
wrong-support escape, where fitting the original points would undo the rescue.
A correction needing more than 3\,cm of collision clearance is rejected.
Otherwise, another drop tests convergence and class-appropriate stability;
failure restores the exact pre-ICP pose.  An already fallen object is not
rejected merely for retaining that same attitude.  Accepted collision shapes
remain in the scene for subsequent objects.  After all objects are placed,
a final joint pass releases every object at once in a single free simulation
run to rest; the resulting small drift is baked into the saved poses and
recorded per object before the scene is imported into Blender.

\subsection{Agentic scene construction}
\label{app:scene-refinement}

\noindent\textbf{Generator--verifier interaction.}
Each stage pairs a VLM generator, which edits and renders the Blender scene,
with a VLM verifier, which compares the render with the reference images.
Both receive the allowed edits for that stage, the object and root-surface
records $\mathcal{O}$ and $\mathcal{R}$, the relation graph $G$, and the current
Blender scene through a stage-entry snapshot and scene-inspection tools.
The snapshot exposes object names, transforms, dimensions, materials, lights,
and camera state; the graph supplies stable IDs, supports, descriptions, and
root-plane evidence.  A stage-specific prompt states which of these quantities
may change.  The original image remains the visual anchor.

\begin{figure}[htbp]
    \centering
    \includegraphics[width=\linewidth]{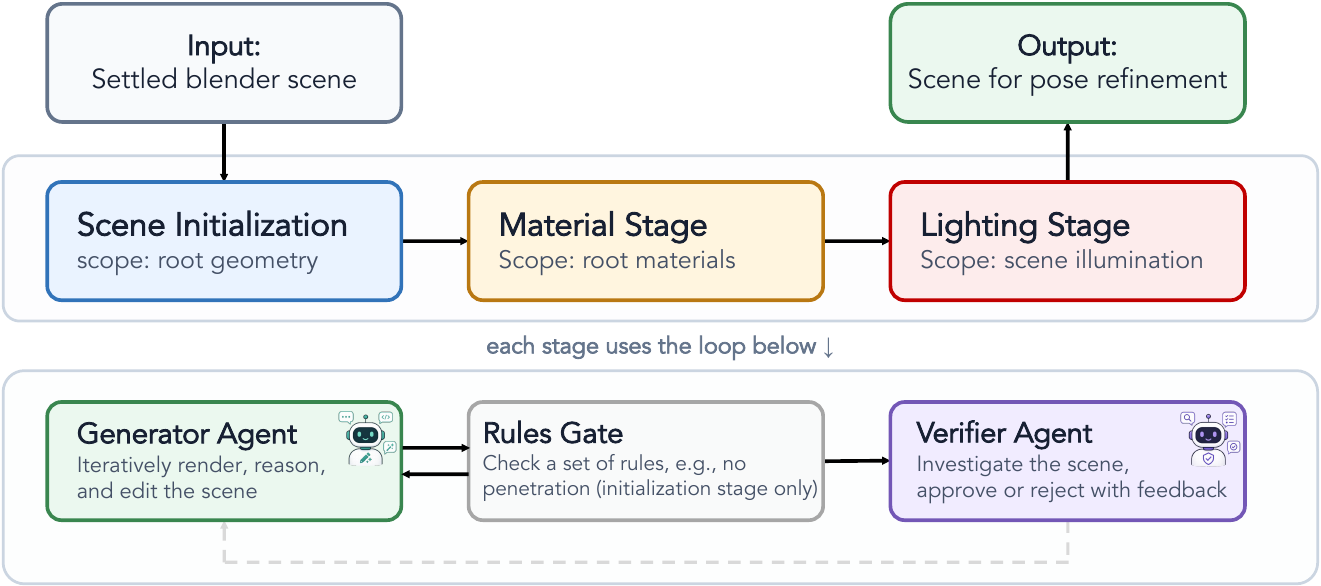}
    \caption{\textbf{Agentic scene construction.} Scene structure, root material, and lighting are refined in a shared Blender scene before object-pose refinement.  Each stage pairs a generator with a verifier. 
    The initialization stage's generator agent additionally has a rules check tool that is required to pass before sending the results to the verifier.
    }
    \label{fig:app-stage-refinement}
\end{figure}

The generator edits scene code, renders the result, and can undo its latest
edit.  The read-only verifier sees a current render and the generator's relevant
actions and rule evidence.  It either approves the stage or returns concrete
corrections within the same scope.  A retry continues from the shared scene,
not a reset scene; initialization also carries forward its construction plan.
Feedback from unrelated stages is not repeatedly inserted into the prompt.
Budgets bound this loop. Reaching a budget without approval produces a
best-effort artifact sent to the subsequent stage. Figure~\ref{fig:app-stage-refinement} shows the shared-scene progression.

\noindent\textbf{Stage-specific edits.}
The initialization generator constructs root geometry and placement from
reliable fitted planes, the reference image, and $E_{\mathrm{surf}}$.
Plane points and normals constrain position and orientation; the photograph
determines shape and finite extent.  A frame-clipped mask supplies only a lower
bound beyond the clipped edge.  The main support is level with its top at
$z=0$, but its horizontal position, footprint, and yaw remain adjustable.
Walls follow their measured horizontal normal and vertical direction through
an orthogonal basis.  Roots lacking reliable planes are
built from their descriptions and the photograph.  A required floor-under-table
relation calls for connected legs or a base reaching the floor, rather than
placing the floor just beneath the tabletop slab.

Every registered root must be represented.  Openings can be cut into a wall,
with small frame or glass details attached to that wall.  A dedicated
root-creation tool can register a genuinely missing architectural wall, and
only such runtime-added roots can be removed through the corresponding tool.
The agent cannot silently delete a preprocessed root or create duplicate
unregistered walls.  One auxiliary level ground is allowed when needed.
Root collision geometry is static, with predefined friction.  Coarse materials
and lighting are initialized once before later appearance refinement.

Imported object poses are the trusted starting point.  Initialization may
translate an object or support stack only through a controlled correction
tool, in response to a measured contact defect or a clear position mismatch
in paired images.  It cannot rotate, resize, remesh, or retexture the object.
The correction tool tests object integrity, support, and new or worsened collisions, rolls back a failed correction, and requires another rule check after success.
Moving an object merely to compensate for a wrongly built support is disallowed.

The appearance stage changes root materials and textures $\tau_r$, including
procedural shader detail, roughness, and bump patterns, while keeping geometry
and object assets fixed.  The illumination stage changes lights, world
illumination, exposure, and view-transform settings, not materials or layout.
It compares only source-view renders with the input image. The source camera stays fixed; alternate views use rendering tools rather than camera edits.

\noindent\textbf{Rule checks and visual feedback.}
As a special tool for the initialization stage only, the read-only \texttt{check\_rules} tool evaluates three levels in sequence.  Structure checks
verify imported-object integrity, registered-root coverage, main-support
geometry, top height, and required structural relations.  Pose checks compare
root silhouettes with the reference masks and verify that all objects rest on their supports. Contact checks detect penetration, floating objects, and violated surface relations.  Advisory relations do not acquire the force of required constraints.  The report identifies the first failing level and its affected entities; later levels are not evaluated and must not be assumed to pass.

The initialization generator revises the scene and reruns the tool after
edits.  Its voluntary completion requires a current pass and verifier approval. For ambiguous or tiny wall masks, an explicit visibility exception can waive an unreliable mask comparison, but not the requirement to build the root. Passing numerical rules establishes a minimum geometric standard, not visual faithfulness: the verifier still judges table edges, wall placement, gaps, and visible structure.  Appearance and lighting have their own visual verifiers without this initialization rule ladder.  Their feedback targets root-surface appearance and source-view lighting, respectively.

\subsection{Physics-Grounded Pose Refinement}
\label{app:pose-refinement}

\noindent\textbf{Object investigation.}
Pose refinement uses one continuous VLM session, without an external verifier or a fresh-memory retry.  
The agent interprets visual evidence and selects tools for geometric measurement, pose search, and simulation. Scene-rendering tools provide source, nearby, and bird's-eye views. 
The investigation tool returns paired, labeled photo/render crops for
one object or a small spatially adjacent group.  Small objects are inspected
alone or in pairs so that their directional features remain visible.
Investigation makes only the listed IDs eligible for bounded moves; the next
investigation replaces that set, while code edits and undo clear it.

The \texttt{Investigate} tool reports silhouette overlap and separate evidence for displacement, projected size, fine yaw, and front--back orientation.  Comparing raw and
centroid-aligned overlap helps detect displacement.  Projected size compares
radial mask extents, or major-axis extents for elongated objects; a large
position error or incompatible aspect ratios suppress an unreliable size
suggestion.  Yaw is measured from silhouette principal axes only when the
axis is sufficiently distinctive.  Missing evidence means unmeasured, not
correct.  The agent checks the paired crops and distinguishes a depth error
from a true size error before resizing.

Front--back evidence compares the current crop and a $180^\circ$ yaw-flipped
render with the photograph using DINOv2~\citep{oquab2024dinov2} patch features.
Weak and strong reversal margins are reported separately; nearly tied or
unavailable features do not establish an orientation.  For elongated objects,
a significant fine-yaw error postpones the flip suggestion until comparable
crops are available.  When an accepted occlusion-completed reference exists,
a known foreground object resting on the target can be hidden in the render
and paired with its removal-edited reference.  Uncertain neighbors and objects
being investigated together remain visible.

\noindent\textbf{Bounded candidate search.}
The agent selects an object and a move aspect $a$ from six choices: horizontal translation along $x$, along $y$, or jointly in $xy$; uniform scaling; yaw adjustment; or a $180^\circ$ yaw flip. The flip follows the special procedure described below. For routine bounded moves, the agent does not specify the adjustment magnitude or search direction.
The \texttt{Move} tool evaluates a coarse candidate set and then finer candidates around the winner. Translation step sizes are proportional to object size, so larger objects are tested with larger shifts. For joint $x$--$y$ moves, the tool compares the rendered object mask with the reference mask and uses their offset to propose a translation. Scale and yaw searches can similarly use measured size and orientation differences to propose candidates. Before rendering each candidate, the tool resolves overlap with the support and adjusts the object's height to place it on the geometry below. These geometric placement operations do not run a physics simulation. Candidates are ranked using the alignment scores below; only the selected candidate proceeds to physical settling and acceptance checks.

\noindent\textbf{Aspect-specific alignment scores.}
Let $m_o$ be the reference mask and $\widehat m_o(X)$ the rendered silhouette
of object $o$ at transform $X$. We write
$\operatorname{IoU}_o(X)=\operatorname{IoU}(\widehat m_o(X),m_o)$;
$\operatorname{cIoU}_o(X)$ aligns the masks' area-weighted centroids before
comparison. Both use coverage-weighted masks, preserving fractional alpha
values and thin structures. The score for aspect $a$ is
\begin{equation}
\label{eq:pose-alignment-score}
A_o^{(a)}(X)=
\begin{cases}
\operatorname{IoU}_o(X)-0.4D_o(X)-0.3Q_o(X), & a=\text{translation},\\
\operatorname{cIoU}_o(X)-0.4D_o(X)-0.3Q_o(X), & a=\text{scale},\\
\operatorname{cIoU}_o(X)+0.2F_o(X), & a=\text{yaw}.
\end{cases}
\end{equation}
Here, $D_o(X)=|d_o(X)-d_o^{\mathrm{ref}}|/|d_o^{\mathrm{ref}}|$ measures
relative depth error. $d_o(X)$ is the depth of the object's center after the candidate translation, and $d_o^{\mathrm{ref}}$ is the median depth of the masked
points of that object. The size penalty
$Q_o(X)=\left|\sqrt{|\widehat m_o(X)|/|m_o|}-1\right|$ compares projected
mask areas. Centering the masks for scale and yaw reduces interference from
translation errors; depth and projected size constrain translation and scale.
Yaw additionally uses appearance to distinguish orientations with similar
silhouettes. Specifically,
\begin{equation}
\label{eq:app-dino-score}
F_o(X)=\frac{1}{|\mathcal{P}_o|}
\sum_{p\in\mathcal{P}_o}
\frac{f_p^{\mathrm{render}}(X)^\top f_p^{\mathrm{photo}}}
{\|f_p^{\mathrm{render}}(X)\|_2\|f_p^{\mathrm{photo}}\|_2},
\end{equation}
where $f_p$ is a DINOv2 ViT-S/14 patch token from an RGB object crop and
$\mathcal{P}_o$ indexes reference-foreground patches in matched-size crops.
If no foreground patch survives downsampling, the full patch grid is used.
The scoring window is the reference
mask's box with $25\%$ padding, and the isolated render is composited on
neutral gray.  
The front--back comparison crops each object to its own box on white and resizes to $224\times224$, reducing sensitivity
to position and scale. Without DINOv2 features, yaw uses silhouette scoring
and supplies no appearance evidence for orientation.

\noindent\textbf{Simulation and acceptance.}
Let $X_o$ be the pre-edit transform, $X'_o$ the proposal tested for acceptance
after any reduction to a feasible fraction, and $\widehat X_o$ its simulated
resting pose. The post-settling score $\widetilde A_o^{(a)}(\widehat X_o)$
recomputes rendered terms while holding the tested proposal's depth penalty
fixed. Let $P_o$ indicate that the applicable proposal-level yaw check and
post-settling physical checks pass. For routine bounded moves, the update is
\begin{equation}
\label{eq:pose-update}
X_o\leftarrow
\begin{cases}
\widehat X_o, &
\begin{aligned}
&A_o^{(a)}(X'_o)-A_o^{(a)}(X_o)\geq\epsilon_a,\\
&P_o=1,\quad
\widetilde A_o^{(a)}(\widehat X_o)\geq A_o^{(a)}(X_o),
\end{aligned}\\[2pt]
X_o, & \text{otherwise}.
\end{cases}
\end{equation}
The aspect-specific threshold $\epsilon_a$ suppresses small proposed gains;
after settling, retaining the baseline score is sufficient.
Table~\ref{tab:app-settings} lists the gain and physical thresholds.
Collision checks compare pre- and post-edit penetration, rejecting
substantial new or worsened overlap; pre-existing overlap is still subject to
the final whole-scene rule gate.  Non-rollable objects are checked for capsize,
while rollable objects are checked for excessive horizontal drift.

A failed candidate can be reduced to a smaller feasible fraction, which must again
clear the gain threshold before simulation. Reliable yaw evidence adds an
improvement test on the selected proposal, with a small noise tolerance.
If appearance features become unavailable during rescoring, their term is
removed from both scores in the post-settling comparison.
Rejection restores all affected transforms, including those of supported objects.

\noindent\textbf{Support propagation and exceptional edits.}
Equation~\eqref{eq:pose-update} applies to routine bounded searches.
The $180^\circ$ flip is a special semantic correction: it need not improve IoU,
but is physically tested and must be followed immediately by a fresh paired
investigation or undo.  A retained or undone applied flip consumes its one-shot
allowance; one physically rejected attempt can be retried.

When moving a support parent, the \texttt{Move} tool compares leaving children in place with carrying them. Children follow if the free-child simulation would topple them or if carrying improves the group's mean overlap.
A resize instead settles the resized parent and then reseats its children and contact-dependent neighbors. 

When bounded moves repeatedly fail or cannot express a correction, the code-edit tool \texttt{Edit} allows direct layout changes, including coordinated edits. The scene-code editing tool detects changed objects, resolves initial overlaps, and settles affected supports before their dependents in simulation. It then returns the post-simulation render and physical diagnostics for the agent to decide whether to retain or undo the simulated result. Automatic alignment and physical acceptance checks govern bounded moves; the agent decides whether to retain simulated code edits. If the required simulation fails to execute, the raw edit is rolled back. Undo restores the pre-edit scene snapshot rather than using approximate inverse transforms.

\begin{figure}[t]
    \centering
    \includegraphics[width=\linewidth]{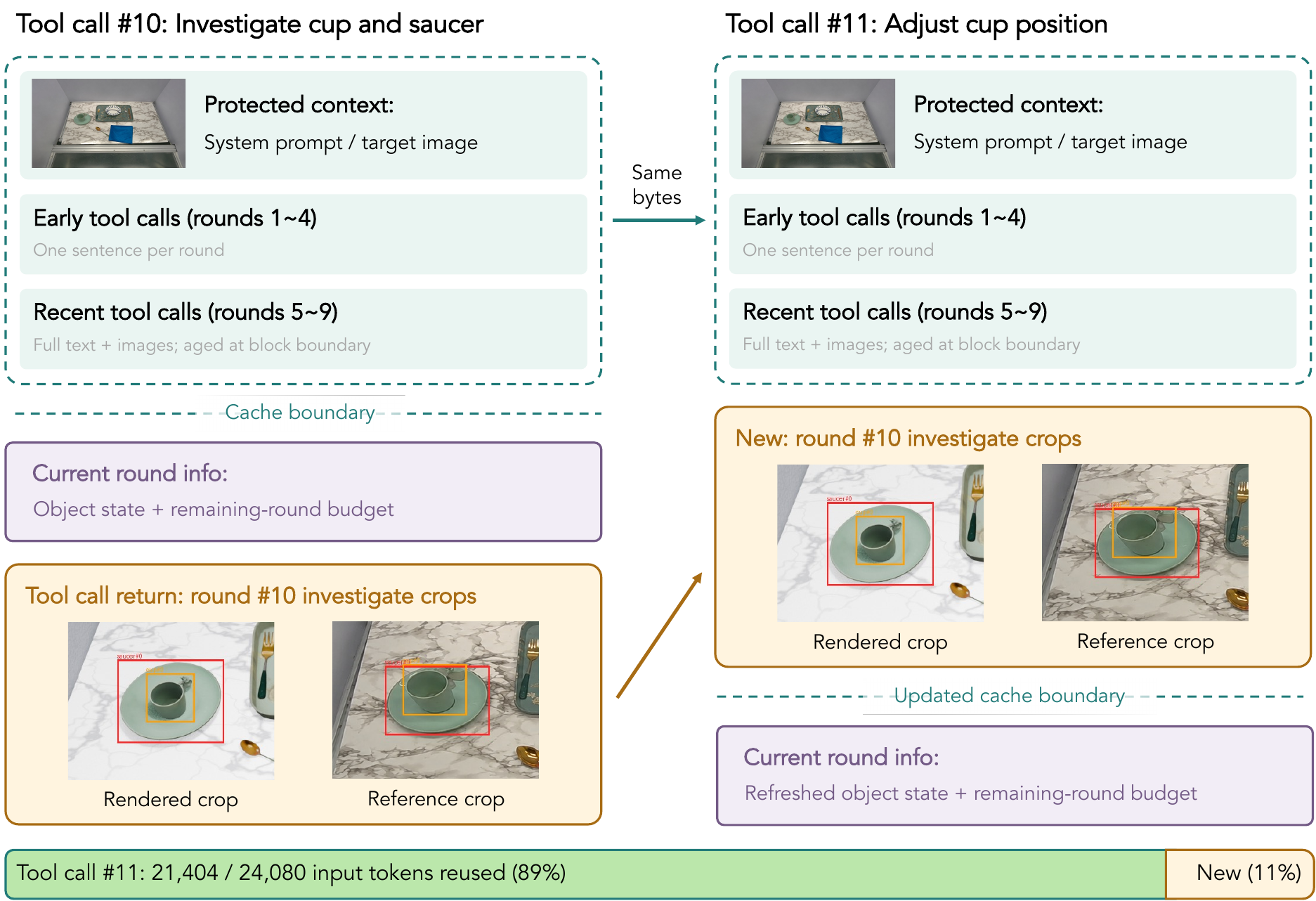}
    \caption{\textbf{Prompt reuse with fresh scene evidence.}
    In this example, tool call~\#11 reuses the unchanged prefix of tool call~\#10 (green), appends the cup/saucer investigation results (amber), and refreshes the object state and budget after the cache boundary (purple). The crops come from the recorded run; the layout illustrates the context policy.}
    \label{fig:app-prompt-caching}
\end{figure}

\noindent\textbf{Context management and prompt caching.}
The agent retains the initial instructions, target image, and stage-entry
scene snapshot.  Recent rounds retain their full text, while
older rounds are compressed in blocks of four into a chronological log,
normally leaving five to eight recent rounds.  The log preserves actions,
scores, acceptance/rejection, and undo outcomes, but removes old images and
stale pose suggestions.  A compression boundary is moved backward if it would
separate an edit from its undo, so the recent window can occasionally grow
beyond eight rounds.

Images are also aged within the visible history.  Superseded full-scene
renders, alternate views, and crop messages become text references to their
saved artifacts; paired crop/reference messages are retained or removed
together.  Duplicate source photographs need not be reattached because the
initial target remains protected.  Aging occurs at the same block boundaries,
not on every round, so it does not continually rewrite the prompt prefix.
The full stored history remains intact; compression changes only what is sent
to the VLM.  The shorter initialization, appearance, and lighting conversations
use append-only history rather than this pose-specific sliding window.

A per-object state table is rebuilt from the full interaction history.  It records the latest measured scores, completed or pending follow-up checks, tool-use counts, rejected aspects, and exhausted budgets. The table and remaining-round budget sit after the cache breakpoint, at the changing end of the prompt.  Their updates therefore preserve the reusable prefix.  Block-aligned compression and image aging allow prompt-cache hits across consecutive calls without freezing the current object state. Figure~\ref{fig:app-prompt-caching} illustrates this mechanism.

Across all 80 benchmark scenes, 84.1\% of pose-refinement input tokens are cache reads, measured over 2,112 logged calls. We divide summed cache-read tokens by total input tokens, including uncached inputs and cache writes in the denominator and excluding output tokens.

\noindent\textbf{Completion and export.}
The completion tool, \texttt{End}, first checks that every retained object has been investigated.  A coverage failure stops this check before the expensive penetration pass, so it does not imply collision clearance.  Once coverage passes, unresolved penetration must be corrected, and any post-flip follow-up must be complete.  Voluntary completion and a current rule pass are both required.  A budget-limited session is explicitly marked incomplete.

After refinement, one joint gravity simulation releases all objects together.
Its resulting transforms are baked into the saved Blender scene; no object
is pinned merely to conceal a final topple.  Non-convergence, large drift, or
toppling is recorded as a warning, and does not by itself establish dynamic
simulation readiness.  Appendix~\ref{app:settings-export} details USD conversion
and its separate readiness check.

\subsection{Implementation settings and simulator export}
\label{app:settings-export}

\noindent\textbf{Model roles and execution settings.}
Table~\ref{tab:app-models} lists the models used in our system and their roles.
In the backbone comparison (Table~\ref{tab:app-backbone-object}), we switch all configurable VLM roles to the selected backbone, including preprocessing, physical-property estimation, scene-construction generators and verifiers, and pose refinement.
Table~\ref{tab:app-settings} reports implementation defaults, which can be overridden by a run configuration.

\begin{table}[h]
    \centering
    \small
    \caption{\textbf{Models and their roles.} The two image-editing paths have
    different inputs and purposes; generated references never replace the real
    source photograph.}
    \label{tab:app-models}
    \begin{tabular}{@{}p{0.35\linewidth}p{0.60\linewidth}@{}}
        \toprule
        Component & Role and conditioning \\
        \midrule
        Configurable VLM & Category/instance proposals, semantic reviews,
        physical-property estimates, and stage agents; launcher default:
        \texttt{claude-opus-5}. \\
        MolmoPoint-8B / SAM3 & Instance-description localization / category,
        synonym, crop, and point-conditioned masks. \\
        MoGe-2 / SAM3D & Metric camera and point map / textured per-object mesh. \\
        LanPaint + Qwen-Image-Edit-2509 & Masked occluder removal and
        one-sided outpainting. \\
        LingBot-Depth & Edited RGB, surviving metric depth, and intrinsics;
        ViT-L/14 v0.5 pretrained checkpoint. \\
        SHARP + GPT-Image-2 & Gaussian novel-view rendering followed by image
        repair; point-splat/masked-edit fallback. \\
        DINOv2 ViT-S/14 & Foreground patch similarity for rotation and
        front--back evidence. \\
        \bottomrule
    \end{tabular}
\end{table}

\begin{table}[t]
    \centering
    \small
    \caption{\textbf{Key implementation defaults.} Physical lengths are in
    meters unless otherwise stated.
    }
    \label{tab:app-settings}
    \begin{tabular}{@{}p{0.27\linewidth}p{0.71\linewidth}@{}}
        \toprule
        Setting & Default \\
        \midrule
        Input / image-edit resolution & RGB input long edge 1536 pixels;
        GPT image-edit input long edge at most 2048 pixels. \\
        Root plane fitting & 200 RANSAC trials, at most 5000 points,
        0.01 inlier distance; reliable root planes require inlier fraction
        $\geq0.6$ and image coverage $\geq2\%$. \\
        Occluder image editing & 20 diffusion steps, guidance 4, seed 0;
        one LanPaint Langevin step per diffusion step. \\
        Ordinary / support CoACD & Concavity 0.05 / 0.02; at most 32 / 64
        parts; a capped support can be retried with 128 parts. Ordinary resolution 50;
        support resolution $\operatorname{clip}(\lceil\ell_{\max}/0.003\rceil,100,200)$,
        where $\ell_{\max}$ is mesh maximum extent. \\
        Physics material fallback & Density 250\,kg/m$^3$; contact offset 0.002, rest offset 0. \\
        Settlement & 120\,Hz; 600-step ordinary or 240-step local budget,
        with a further 600 steps for an unconverged drop. \\
        Rest detection & After at least 30 steps, 20 consecutive steps with
        translation $<5\times10^{-5}$ m and rotation $<0.0005$ radians per step. \\
        ICP & At most 60 iterations, $\pm30^\circ$ yaw, scale 0.85--1.18,
        at least 150 target points; scale locked for equal-size groups. \\
        Move coarse probes & Axis translation $\pm\{0.05,0.15,0.4\}$ times
        object size; yaw $\pm\{5,15,40\}^\circ$; scale
        $\{0.8,0.9,1.1,1.25\}$, plus measured proposals and fine probes. \\
        Move gain thresholds & Translation 0.01; scale 0.01, or 0.005 with
        an active size hint; centered rotation score 0.012. \\
        Move stability tests & Ordinary capsize threshold $45^\circ$;
        rollable horizontal drift 50\,mm.  Penetration fails when depth
        exceeds 8\,mm and worsens by more than 5\,mm. \\
        Agent round budgets & Initialization 20, appearance 10, lighting 10;
        verifiers 8 each; up to 2 attempts per paired stage.  Pose refinement:
        one session of $\min(70,8+5|\mathcal{O}|)$ rounds. \\
        Per-object pose budgets & 3 investigations and 5 moves;
        code relocations have a soft cap of 3. \\
        \bottomrule
    \end{tabular}
    \vspace{-4mm}
\end{table}


\noindent\textbf{Simulator Export.} USD conversion is a separate step after Blender reconstruction.  It bakes procedural materials and transfers meshes, textures, lights, and camera state,
then assigns collision geometry and physical properties.  An identity
map links each pipeline object to its USD primitive, accounting for name
sanitization.  Missing, ambiguous, or nested object identities fail conversion
rather than losing bodies.  Mass and inertia are adjusted consistently for
the exported geometry.  A simulator-side verification checks the exact object
inventory, dynamic versus kinematic state, support contact, convergence,
translation and rotation drift, and final velocities. Reconstruction completion,
conversion success, and dynamic readiness are recorded separately.

\section{Additional Experimental Details}
\label{app:experimental-details}


\subsection{Baseline implementation details}
\label{app:baseline-implementation}

We adapt the baseline implementations to the same single-image benchmark.
Each method retains its reconstruction pipeline and scene representation;
method-specific adapters expose the reconstructed objects and evaluation
renders.  Below we describe these adaptations and, in particular, how a
source camera and a nearby evaluation camera are obtained.

\noindent\textbf{VIGA.}
We use VIGA's static-scene configuration with SAM-based segmentation and
SAM3D asset initialization, followed by iterative Blender code generation,
rendering, and verifier feedback.  To support object-centric evaluation and simulation, we augment
the generator's system prompt with a simple object-grouping contract
(one parent per movable object with every part
beneath it, one parent for the static environment, no loose top-level meshes),
and after every code execution the tool result lists the meshes that are not
yet grouped, so the agent can repair them before it stops.  We use a 20-round
generator budget with medium reasoning effort for every backbone; the same VLM
also names the segmented objects.  Grouping changes neither the renderer nor
the assets, and the agent complied in every scene.
Intermediate renders preserve the input aspect ratio with a 512-pixel long
edge.  For evaluation, we select the latest numbered render directory containing
a rendered frame and load its saved Blender scene.  We select the main camera
by name, falling back to the camera named \texttt{Camera} and then an available
camera, and retain its saved pose and intrinsics.  This avoids using a
diagnostic camera left active.  Evaluation renders use the
input-image dimensions.

\noindent\textbf{SceneConductor.}
We run the three-stage pipeline: mask review and merging, SAM3D asset
generation and GALP layout initialization; environment construction; and
relation-guided scene refinement, including refinement of selected object
groups in isolated subscenes.  Its source camera uses GALP's predicted focal
length and orientation in the camera-centered reconstruction frame, with the
pipeline's coordinate-convention correction.  We render the final scene at
the input-image dimensions, preserving its aspect ratio instead of using the
default fixed-size diagnostic views.  Walls that obstruct the evaluation
camera are hidden from camera rays.  The nearby-view adapter also uses a gray
world background with unit-strength ambient illumination.

\noindent\textbf{SimFoundry.}
We use the pipeline's single-image mode: the input is supplied as one RGB
frame, with a repeated-frame video only to satisfy the video input interface.
Reconstruction uses DA3 depth and camera estimates, support-plane alignment,
object extraction and image cleanup, Hunyuan3D asset generation,
FoundationPose refinement, and physics preparation and settling before
OmniGibson scene assembly.  The source camera combines the estimated intrinsics
with the camera-to-world transform from support-plane alignment.  The
single-image configuration uses the simulator's planar support surface.
Pairwise evaluation renders use OmniGibson at a 1536-pixel long edge with the
input aspect ratio; the simulator's added robot is hidden.

\noindent\textbf{REST3D and source-camera recovery.}
We evaluate REST3D's final object reconstruction after its scene assembly and
stabilization stages.  Its exported scene does not retain a calibrated source
camera, so our evaluation adapter recovers an approximate camera from the
input image and the reconstruction intermediates.  We rerun the same
MoGe model used by its asset pipeline to estimate image intrinsics.
To recover the pose, we fit the reconstructed raw floor/carpet geometry with
a plane, use its normal to recover the pipeline's up-axis alignment, and
propagate the raw camera origin, forward direction, and up direction through
that alignment.  We then apply the support-plane height shift and the
conversion from the pipeline's $Y$-up frame to the evaluation $Z$-up frame.
This recovery is approximate because REST3D's subsequent per-object grounding
and stabilization are not a single rigid scene transform.  The camera is
derived geometrically from these intermediates, without selecting it by
render similarity or using a ground-truth simulator camera.
The final meshes are converted to USD with vertex colors retained and rendered
in Isaac Sim at the input resolution, using the recovered horizontal focal
length, square pixels, and a centered principal point.  A synthetic support
slab is added because the raw floor reference is omitted from the final object
export.  For the shared stability test, REST3D's reconstructed tables and desks and its one wall-mounted object are static colliders; every other REST3D mesh is a dynamic rigid body on the support slab.

\noindent\textbf{SAM3D object baseline.}
This comparison shares \method's segmentation and generated object assets.
We retain SAM3D's predicted rotation, translation, and scale, applying only
the camera-coordinate conversion and the shared gravity alignment to place
them in the scene frame.  We bypass \method's depth-based recentering and
rescaling, registration, settling, and subsequent object-pose refinement.
For rendering, these raw-pose assets replace the objects in \method's final
scene while retaining its source camera, support surfaces, walls, and lighting.
Thus this is an object-reconstruction and placement baseline with shared
scene context.

\noindent\textbf{Source and nearby view selection.}
Source-view metrics use the camera supplied or recovered for each method as
described above.  For pairwise judging, we render a nearby view
by rotating the source camera position $c$ by $+30^\circ$ around a pivot $p$
and the reconstruction's vertical axis:
\begin{equation}
    c_{\mathrm{near}} = p + R_z(30^\circ)(c-p).
\end{equation}
The focal length and the camera's height and distance relative to $p$ are preserved. Each nearby view is thus a corresponding viewpoint change in the reconstruction's own frame, not a shared ground-truth extrinsic, and is used only for evaluation.

The pivot $p$ is where the source optical axis meets a horizontal reference plane. REST3D and SimFoundry use the support plane $z=0$; SceneConductor uses its estimated floor height; VIGA uses the median height of visible mesh bounding-box centers. If the intersection is invalid, REST3D falls back to its scene bounding-box center, SceneConductor to the placed-object centroid, and VIGA to the component-wise median of the visible mesh bounding-box centers (also used when the intersection lies too far outside the scene); the REST3D and SceneConductor fallbacks are projected onto their reference planes. Because each camera is aimed at its pivot, a fixed orbital elevation does not guarantee identical roll, or identical pitch when a fallback pivot is used.

The source and nearby images are presented to the judge at common tile sizes,
preserving aspect ratio, together with the original input photograph
(Appendix~\ref{app:vlm-physics-protocol}).  Rendering remains method-specific:
VIGA and SceneConductor use Blender/Cycles, REST3D uses Isaac Sim, and
SimFoundry uses OmniGibson.  The scored REST3D and SimFoundry renders use their
brown synthetic support surfaces.  

\subsection{Benchmark and metric details}
\label{app:evaluation-metrics}

\noindent\textbf{Benchmark composition.}
The 80-scene benchmark comprises 44 simulator-grounded scenes and 36 additional scenes whose input images consist of 18 real robot workspace images, 6 phone captures, 8 internet images, and 4 AI-generated images.

\noindent\textbf{Image and instance metrics.}
Full-frame metrics compare the source-camera render with the input image.
LPIPS uses AlexNet features; nCLIP is one minus the cosine similarity of CLIP
ViT-B/32 image embeddings.  The reported PSNR uses raw RGB errors, without
fitting an exposure or color correction.  Object-only PSNR pools squared errors over ground-truth object pixels, excluding the support table.  Object-only SSIM averages the spatial SSIM map over the same pixels.

For object-centric LPIPS and nCLIP, ground-truth instance masks define crop
boxes shared by the reference and render.  Boxes have $10\%$ padding and
a minimum side of 64 pixels; objects with fewer than 200 visible pixels are
excluded from crop metrics.  Within each box, pixels outside the object's
ground-truth mask are set to white in both images, so these metrics score the
object without its surroundings.  Crop scores are averaged
over objects within a scene.

Ground-truth mesh parts are grouped into logical objects before instance
evaluation.  ARI compares the two instance partitions over all pixels, with
background, including the table, treated as one cluster.  FG-IoU is the
intersection over union between the predicted and ground-truth object
foregrounds, each the union of its object pixels with the table excluded.
Thus ARI measures partitioning while FG-IoU measures detection coverage:
missed, hallucinated, or misplaced objects lower FG-IoU even when the
remaining pixels are partitioned consistently.  Count MAE is the mean absolute difference between the predicted and ground-truth object counts in the evaluated instance masks. For final-scene evaluation, the predicted count includes only objects visible in the source-view render, rather than all objects stored in the reconstructed scene.

\noindent\textbf{Canonical 3D layout evaluation.}
Ground-truth and predicted positions are expressed in corresponding canonical frames whose origins are the source viewing axis’s intersections with the support plane. The vertical axis is gravity-up, and the horizontal orientation is fixed by the source-camera direction. If either camera cannot define this intersection, each set instead uses the centroid of all objects' positions projected onto the support plane. For both predictions and ground truth, object positions are defined by world-space axis-aligned bounding-box centers and then transformed into the canonical frame.
All predicted objects contribute to the metric, including unmatched extras. We evaluate object arrangement up to a global scale factor, allowing comparison across reconstruction methods with different scale conventions.

We fit only a global scale.  Candidate scales include one and
ratios of the two sets' pairwise-distance quantiles at $25$, $50$, $75$, and
$90\%$.  Hungarian matching at a 20\,cm cutoff ranks candidates first by match
count and then by mean matched distance.  Up to two least-squares refinements
use $s=\sum_i p_i^\top g_i/\sum_i\|p_i\|_2^2$ for matched predicted and true
positions $(p_i,g_i)$.  Refinement requires at least three matches and cannot
reduce their count.  This scale is shared across all reported cutoffs.

For point sets $P$ and $Q$ with $m\leq n$ elements and $n>0$, OSPA of order one is
\begin{equation}
\label{eq:app-ospa}
\operatorname{OSPA}_c(P,Q)=\frac{1}{n}\left[
\min_\pi\sum_{i=1}^{m}\min\!\left(c,\|p_i-q_{\pi(i)}\|_2\right)
+c(n-m)\right],
\end{equation}
where $\pi$ assigns the smaller set to distinct members of the larger.
Two empty sets have zero distance.
It penalizes localization error and unmatched objects
\citep{schuhmacher2008ospa}.  We report 5 and 10 cm cutoffs and
F1 from a separate 10\,cm-threshold assignment on these aligned positions.

\subsection{Pairwise VLM and physics protocols}
\label{app:vlm-physics-protocol}

For pairwise visual evaluation, GPT-5.6-terra and
Qwen3.5-122B-A10B~\citep{qwen35} compare reconstruction pairs for completeness,
layout, and appearance.  The nearby evaluation view uses the $30^\circ$ orbit
and method-specific camera recovery described in
Appendix~\ref{app:baseline-implementation}, keeping the camera's orbital
elevation and distance fixed.  Each candidate panel shows its source and nearby render,
with anonymous labels and uniform tile sizes; the judge also sees the real
input photograph.  Each pair is presented in both candidate orders.
We remove order-inconsistent judgments, give ties half credit, and fit
Bradley--Terry scores to summarize pairwise preferences.  Half a pseudo-win is
added in each direction for every compared method pair to avoid infinite
strength estimates for undefeated methods.  Scores are centered
to have zero mean across methods for each criterion.
After excluding comparisons for which either judge is order-inconsistent, the two judges achieve Cohen’s $\kappa=0.87$. Order-inconsistent verdicts, discarded as above, make up 13--23\% of paired judgments per criterion.

Physical stability is evaluated separately by running each reconstructed scene for 300 PhysX steps at 60\,Hz.  Displacement is the Euclidean distance between an object's initial and final position, not accumulated travel along its path.
An object passes the stability test if its displacement after 5 s is below 50 mm and its final velocity is zero; a scene passes only if every object passes.
The scene-level rate is the fraction of passing scenes, while the object-level rate pools passing objects across scenes.  

\subsection{Real-to-sim policy evaluation protocol}
\label{app:sim2real-protocol}

This section details the two policy evaluations in
Section~\ref{sec:sim2real-eval}. Table~\ref{tab:sim2real} reports $\pi_{0.5}$ and Table~\ref{tab:lingbot-sim2real} reports LingBot-VLA 2.0; both policies are fine-tuned on the same in-house teleoperation data and evaluated on separate sets of 50 real episodes across the same five tasks: ten episodes per policy--task pair, totaling 100 real episodes. Each episode is evaluated in both reconstruction methods with its corresponding policy checkpoint. The five tasks evaluated are: everything into the bin, fruit on plate, fruits from plate to bowl, marker in cup, and mustard in bin.

\textbf{Scene export.} Each scene is reconstructed by \method (or baseline) from an armless head-camera image and exported to Isaac Sim together with the calibrated camera rig of the station, the episode's initial arm configuration (reset pose), and the episode's text instruction. Unlike the RGB-only benchmark of Section~\ref{sec:experiments}, these station scenes provide metric depth: both \method and SimFoundry receive the head camera's color-aligned depth map, hole-completed with LingBot-Depth~\citep{tan2026lingbotdepth}, together with the calibrated intrinsics, and use it in place of monocular depth estimation. Both methods use their own mass estimates for objects, and the friction coefficient is derived from the estimated material; contacts use the larger friction coefficient of the two bodies. 
The gripper fingertips use signed-distance-field collision shapes with a friction coefficient of 8 (static) and 7 (dynamic). We choose these friction values following manipulation simulation benchmarks such as RoboDojo~\citep{chen2026robodojo} and RoboLab~\citep{yang2026robolab}, so that objects can be reliably grasped in simulation.

\textbf{Closed-loop rollout.} Physics is stepped at 240\,Hz, and control runs
at 30\,Hz.  Each action chunk is executed for
50 control steps (1.67\,s) before the next policy query; the episode is
therefore closed loop, unlike the open-loop trajectory replay in
Appendix~\ref{app:trajectory-replay-protocol}.  An episode ends
after 360\,s, once both arms have been stationary for 10\,s (fingertips and
gripper joints moving less than 5\,mm), or once no object has moved more than
5\,mm for 60\,s, counted from the first object motion.  The same settings are
used for the \method and the SimFoundry rows.

\textbf{Scoring.} Simulated success is a task-specific geometric predicate on the final object centroids relative to the container at its final position. Plates and bowls use a circle fitted to the outer rim, cups and mugs a circle fitted to the wall (which excludes a mug's handle), and bins the axis-aligned box of the bin mesh. Required objects are defined by the real scene: the fruit tasks require every fruit, mustard in bin requires the mustard bottle, and everything into the bin requires every object that did not start inside the bin. A required object missing from the reconstruction counts as a failure, and for everything into the bin, hallucinated objects must also be placed in the bin. A human scorer verifies the verdicts generated by the script. Real success is the operator's verdict.

\textbf{Metric aggregation.} For each policy, agreement and Cohen's $\kappa$ use its 50 episodes, and Pearson's $r$ correlates its five real and simulated task success rates. Combined agreement pools all 100 episodes. The pooled correlation is computed directly from all ten policy--task success-rate pairs. MMRV~\citep{li2024simpler} averages, over policy--task pairs, the largest real success-rate gap to any pair whose simulated order contradicts the real order; lower is better. Confidence intervals come from 10{,}000 paired bootstrap resamples of episodes within each policy--task cell, and McNemar's exact test compares the two methods' per-episode agreement with the real verdicts. With five tasks per policy, the rate-level metrics ($r$, MMRV, and rate error) are reported without intervals.

\noindent\textbf{MMRV.}
We follow SIMPLER's published formula~\citep{li2024simpler}, using strict $<$ comparisons. For $K$ paired real and simulated success rates $r_i$ and $s_i$,
\[
V_{ij}=|r_i-r_j|\,\mathbf{1}\!\left[(s_i<s_j)\ne(r_i<r_j)\right],
\qquad
\mathrm{MMRV}=\frac{1}{K}\sum_{i=1}^{K}\max_{1\le j\le K}V_{ij}.
\]
Across the ten pooled policy--task pairs, MMRV is 0.04 for \method and 0.11 for SimFoundry (Table~\ref{tab:robotics-stats}); lower is better.

\textbf{SimFoundry inference.} For the SimFoundry rows, the reconstructed
objects are replaced by SimFoundry's meshes, textures, poses, and masses,
and registered into the station frame through the shared head camera. 
Since SimFoundry does not reconstruct table mesh, we borrow it from our method and paint the table with SimFoundry's inpainted table image.

\begin{table}[t]
    \centering
    \footnotesize
    \setlength{\tabcolsep}{3pt}
    \captionsetup{font=small}
    \caption{\textbf{Real-to-sim evaluation with LingBot-VLA 2.0.}
    The same fine-tuned checkpoint is evaluated on 50 real episodes and both
    reconstructions. Success counts successful episodes; agreement counts
    matching real and simulated verdicts. Cohen's $\kappa$ uses all 50 episodes;
    Pearson's $r$ uses the five task-level success rates.}
    \label{tab:lingbot-sim2real}
    \vspace{-2mm}
    \resizebox{\linewidth}{!}{%
    \begin{tabular}{p{2cm} l c M{2.0cm} M{2.0cm} M{2.0cm} M{2.0cm} M{2.0cm}|cc}
        \toprule
        Environment & & Everything$\to$bin & Fruits$\to$plate & Fruits$\to$bowl & Marker$\to$cup & Mustard$\to$bin & All & Cohen's $\kappa\uparrow$ & Pearson $r\uparrow$ \\
        \midrule
        Real-world & success & 5/10 & 7/10 & 2/10 & 3/10 & 3/10 & 20/50 & -- & -- \\
        \midrule
        \multirow{2}{*}{SimFoundry} & success & 4/10 & 4/10 & 0/10 & 0/10 & 2/10 & 10/50 & -- & 0.88 \\
        & agreement & 5/10 & 5/10 & \textbf{8/10} & \textbf{7/10} & 5/10 & 30/50 & 0.09 & -- \\
        \midrule
        \multirow{2}{*}{\textbf{\method}} & success & 5/10 & 6/10 & 1/10 & 1/10 & 0/10 & 13/50 & -- & \textbf{0.93} \\
        & agreement & \textbf{6/10} & \textbf{7/10} & 7/10 & 6/10 & \textbf{7/10} & \textbf{33/50} & \textbf{0.25} & -- \\
        \bottomrule
    \end{tabular}}
    \vspace{-1mm}
\end{table}

\textbf{Agreement across policies.}
For LingBot-VLA 2.0, \method achieves 66\% episode agreement ($\kappa=0.25$)
versus 60\% ($\kappa=0.09$) for SimFoundry, with task-level correlations of
$r=0.93$ and $r=0.88$, respectively. Both underestimate real-world success:
\method succeeds on 13/50 episodes and SimFoundry on 10/50, versus 20/50
on the real robot. The agreement advantage varies by task; SimFoundry agrees
more often on fruits-to-bowl and marker-to-cup. Across both policies,
\method matches 69/100 real outcomes, versus 54/100 for SimFoundry.
Figure~\ref{fig:policy-correlation} combines their ten policy--task
success-rate pairs, yielding pooled correlations of $0.92$ and $0.84$.
These correlations capture variation across both policies and tasks;
the tables report each policy separately.

\textbf{Statistical analysis.} Table~\ref{tab:robotics-stats} adds uncertainty estimates and paired tests. Pooled over both policies, \method's agreement advantage is significant (25 vs.\ 10 discordant episodes, McNemar $p=0.017$), and its $\kappa$ exceeds SimFoundry's by 0.29 (95\% CI [0.08, 0.50]). 

\begin{table}[t]
    \centering
    \footnotesize
    \setlength{\tabcolsep}{3pt}
    \captionsetup{font=small}
    \caption{\textbf{Statistical analysis of real-to-sim agreement.} Agreement and Cohen's $\kappa$ use per-episode verdicts, with 95\% paired-bootstrap intervals. McNemar counts the episodes on which only \method or only SimFoundry agrees with the real verdict. Pearson's $r$, MMRV, and rate error (the mean absolute difference between real and simulated task success rates) use task-level success rates and are descriptive.}
    \label{tab:robotics-stats}
    \vspace{-2mm}
    \resizebox{\linewidth}{!}{%
    \begin{tabular}{llccccc|c}
        \toprule
        Policy & Method & Agreement $\uparrow$ & Cohen's $\kappa\uparrow$ & Pearson $r\uparrow$ & MMRV $\downarrow$ & Rate error $\downarrow$ & McNemar \\
        \midrule
        \multirow{2}{*}{$\pi_{0.5}$} & SimFoundry & 48\% [36, 62] & $-0.07$ [$-0.32$, 0.18] & $-0.38$ & 0.12 & 0.16 & \multirow{2}{*}{18 vs.\ 6, $p=0.023$} \\
        & \method & \textbf{72\%} [60, 84] & \textbf{0.38} [0.11, 0.63] & \textbf{0.70} & \textbf{0.00} & \textbf{0.12} & \\
        \midrule
        \multirow{2}{*}{LingBot-VLA 2.0} & SimFoundry & 60\% [46, 74] & 0.09 [$-0.16$, 0.35] & 0.88 & 0.06 & 0.20 & \multirow{2}{*}{7 vs.\ 4, $p=0.55$} \\
        & \method & \textbf{66\%} [52, 78] & \textbf{0.25} [0.00, 0.50] & \textbf{0.93} & \textbf{0.04} & \textbf{0.14} & \\
        \midrule
        \multirow{2}{*}{Pooled} & SimFoundry & 54\% [45, 63] & 0.09 [$-0.08$, 0.27] & 0.84 & 0.11 & 0.18 & \multirow{2}{*}{25 vs.\ 10, $p=0.017$} \\
        & \method & \textbf{69\%} [60, 78] & \textbf{0.38} [0.20, 0.56] & \textbf{0.92} & \textbf{0.04} & \textbf{0.13} & \\
        \bottomrule
    \end{tabular}}
    \vspace{-1mm}
\end{table}

\subsection{Open-loop trajectory replay protocol}
\label{app:trajectory-replay-protocol}

This section details the open-loop trajectory replay experiment in
Section~\ref{sec:trajectory-replay}.  Unlike the evaluation in
Section~\ref{sec:sim2real-eval}, which queries the policy between action chunks,
this experiment executes an entire command sequence recorded on the real robot
without additional policy inference.

\textbf{Episodes.} We evaluate 25 successful teleoperated demonstrations recorded
on a bimanual station of the same design, with arm-base and head-camera poses
calibrated using a ChArUco board before recording. There are five demonstrations
per task: everything into the bin, cup into the bowl, fruits from the plate into
the bowl, marker into the cup, and mustard bottle into the bin. The demonstrations
involve 1--6 evaluated objects each. We replay the recorded follower's joint
trajectory, logged at 30--60\,Hz depending on the session, rather than the leader
commands, so that the simulated arm follows the motion observed in the recording.

\textbf{Scene construction and registration.} Both methods receive the same armless head-camera image and metric depth input as in Appendix~\ref{app:sim2real-protocol}. For each episode, we import the
reconstructed textured geometry and estimated physical properties into Isaac
Sim. We align the reconstructed scene to the station frame using the calibrated head-camera pose.
The simulated rig uses the measured robot-base poses and the camera's
measured intrinsics.  Collision geometry is derived from the reconstructed
object meshes, the tabletop is a static prism collider at the measured height,
and the robot is initialized from the episode's recorded joint configuration.
For SimFoundry, we register its objects into the same rig frame through the
head camera, as in the object-replacement protocol above, keep its estimated object
masses, assign friction from our material table by object category, and replay
the same command stream in the same calibrated robot setup.  SimFoundry scenes
are rendered with its inpainted table texture and \method scenes with the
reconstruction's own tabletop.

\textbf{Command replay.} Physics is stepped at 240\,Hz.  After a 0.5\,s settling period, we execute the time-stamped joint targets and gripper commands through the simulated controllers according to their recorded timing. The arm controller uses hardware-matched joint gains with full gravity feed-forward.
After initialization, the robot follows the commands through simulated controllers, while objects move through the resulting contacts and dynamics. No policy queries, trajectory adaptation, or human intervention occur after replay begins. 

\textbf{Scoring.} We score final object placements with the criteria in Appendix~\ref{app:sim2real-protocol}, including treating a missing required object as a failure, with automated verdicts reviewed against the real demonstrations by a human scorer. For everything-into-the-bin replay, objects already inside the bin are excluded from the placement count. We also exclude objects that the real demonstration leaves untouched. These exclusions apply to both methods; hallucinated objects are not exempt, so an episode fails if one remains outside the bin.

Replays succeed in 20 of 25 \method scenes and 9 of 25 SimFoundry scenes (95\% Wilson intervals [61, 91]\% and [20, 55]\%). \method succeeds on all nine demonstrations that SimFoundry replays successfully and on eleven more (exact McNemar $p=0.001$).
Across the \method scenes, 39 of 45 evaluated objects reach their targets.
Of the five unsuccessful replays, four leave one object unplaced and one leaves
two. SimFoundry's automatic object counts include extra instances and
can omit missing or fused objects, so they do not provide a directly comparable
object-level success fraction. We therefore compare the methods using the
reviewed episode-level outcomes in Table~\ref{tab:replay}.

\noindent\textbf{Placement ablation.}
We repeat the same 25 demonstrations with the code-edits-only variant, which
removes the bounded move tool while retaining direct scene-code edits and
physical settling. Recorded commands, calibration, controllers, simulation
settings, and scoring remain unchanged. This variant succeeds on 11/25
replays (44\%), compared with 20/25 (80\%) for full \method
(Table~\ref{tab:replay-placement-ablation}). The full method improves success
on four tasks and ties on mustard into the bin, supporting the value of the
bounded move tool for placement fidelity in robot manipulation.

\begin{table}[htbp]
    \centering
    \small
    \captionsetup{font={small}}
    \caption{\textbf{Object placement ablation for open-loop replay.}
    Successful replays out of five real demonstrations per task. Code edits
    only removes \texttt{Move} while retaining \texttt{Edit}; both variants use the same replay protocol.}
    \label{tab:replay-placement-ablation}
    \setlength{\tabcolsep}{12pt}
    \begin{tabular}{lcc}
        \toprule
        Task & Code edits only & Full \method \\
        \midrule
        Everything$\to$bin & 1/5 & \textbf{3/5} \\
        Cup$\to$bowl & 1/5 & \textbf{4/5} \\
        Fruits$\to$bowl & 2/5 & \textbf{4/5} \\
        Marker$\to$cup & 2/5 & \textbf{4/5} \\
        Mustard$\to$bin & \textbf{5/5} & \textbf{5/5} \\
        \midrule
        All & 11/25 (44\%) & \textbf{20/25 (80\%)} \\
        \bottomrule
    \end{tabular}
\end{table}

\section{Additional Experiments}
\label{app:additional-exp}

\subsection{Ablation Studies}
\label{app:ablation}

\begin{figure}[t]
    \centering
    \includegraphics[width=\linewidth]{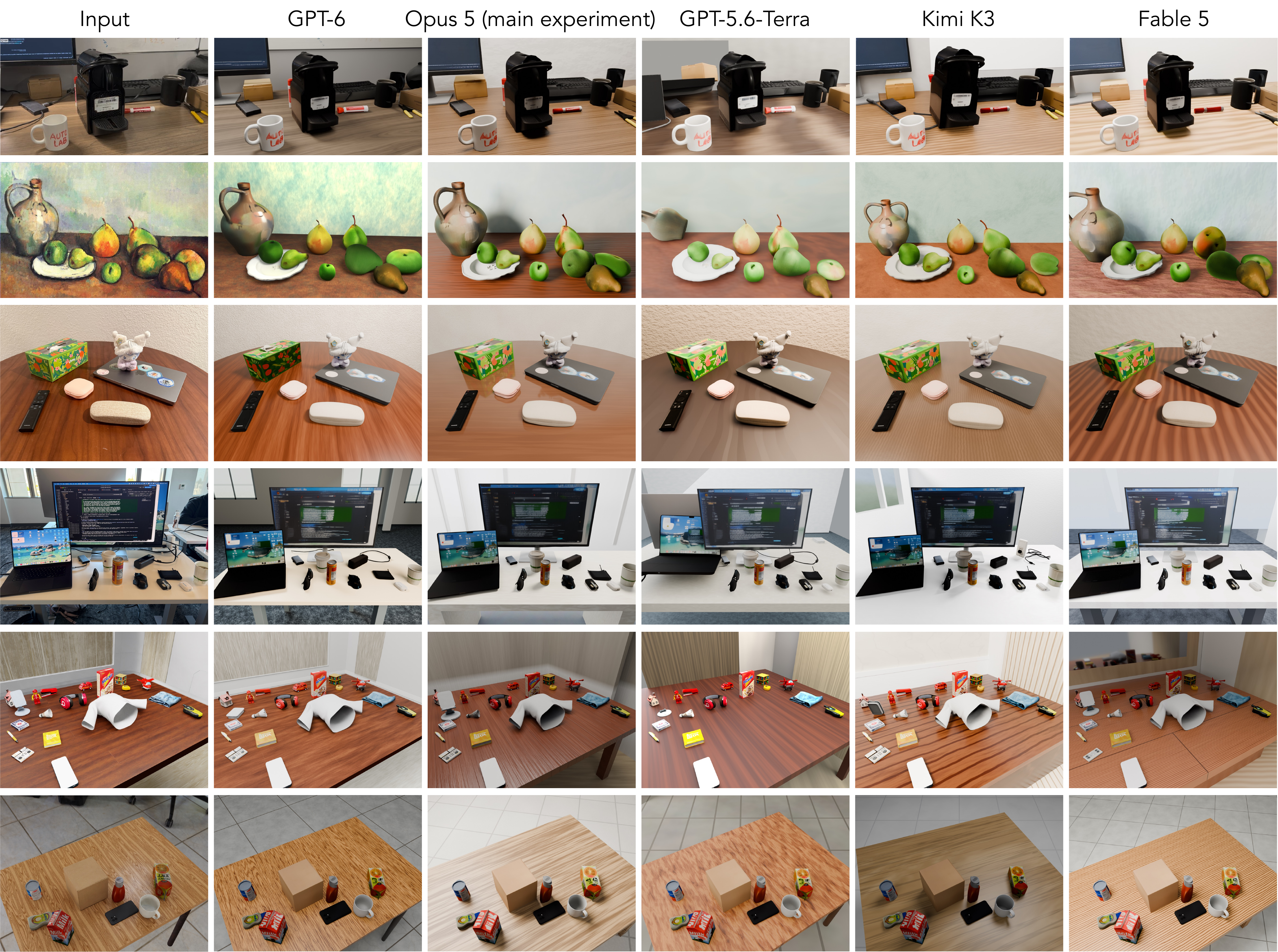}
    \caption{\textbf{Qualitative comparison across VLM backbones.}}
    \label{fig:app-vlm-backbones}
\end{figure}

\noindent\textbf{Backbone comparison.}
Table~\ref{tab:app-backbone-source} evaluates \method with different VLM
backbones (main experiments use Claude Opus 5). GPT-6 Astra achieves the best
results across all reported metrics, with the clearest gains in appearance.
Object placement remains similar across backbones, suggesting that improved
rendered similarity does not necessarily translate into comparable gains in
3D layout.  Figure~\ref{fig:app-vlm-backbones} provides qualitative comparisons.

\begin{table}[h]
    \centering
    \small
    \caption{\textbf{Reconstruction with different VLM backbones.} Every row uses \method with the listed backbone for all configurable VLM roles across preprocessing and the agent stages.}
    \label{tab:app-backbone-source}
    \label{tab:app-backbone-object}
    \vspace{-2mm}
    \setlength{\tabcolsep}{3pt}
    \resizebox{\linewidth}{!}{%
    \begin{tabular}{p{2.4cm} |cccc|cccc|cc|cc}
        \toprule
        & \multicolumn{4}{c}{Source-view reconstruction}
        & \multicolumn{4}{c}{Object-centric appearance}
        & \multicolumn{2}{c}{Instance structure}
        & \multicolumn{2}{c}{3D object layout} \\
        \cmidrule(lr){2-5} \cmidrule(lr){6-9} \cmidrule(lr){10-11} \cmidrule(lr){12-13}
        VLM backbone & PSNR $\uparrow$ & SSIM $\uparrow$
        & LPIPS $\downarrow$ & nCLIP $\downarrow$
        & PSNR$^\ast$ $\uparrow$ & SSIM$^\ast$ $\uparrow$
        & LPIPS$^\ast$ $\downarrow$ & nCLIP$^\ast$ $\downarrow$
        & ARI $\uparrow$ & FG-IoU $\uparrow$
        & OSPA@5 $\downarrow$ & F1@10 $\uparrow$ \\
        \midrule
        Claude Opus 5 & 13.277 & 0.544 & 0.518 & 0.122
        & 13.609 & 0.440 & 0.170 & 0.062 & 0.878 & 0.819 & 2.4 & 0.93 \\
        Claude Fable 5 & 12.837 & 0.535 & 0.535 & 0.126
        & 13.082 & 0.432 & 0.170 & 0.064 & 0.879 & 0.820 & 2.4 & 0.94 \\
        GPT-5.6-terra & 13.918 & 0.577 & 0.527 & 0.128
        & 13.338 & 0.437 & 0.178 & 0.067 & 0.850 & 0.781 & 2.6 & 0.92 \\
        Kimi K3 & 12.830 & 0.542 & 0.557 & 0.121
        & 13.196 & 0.439 & 0.184 & 0.066 & 0.865 & 0.802 & 2.5 & 0.93 \\
        GPT-6 Astra & \textbf{16.512} & \textbf{0.592} & \textbf{0.369} & \textbf{0.100}
        & \textbf{14.535} & \textbf{0.462} & \textbf{0.151} & \textbf{0.060}
        & \textbf{0.891} & \textbf{0.838} & \textbf{2.3} & \textbf{0.95} \\
        \bottomrule
    \end{tabular}}
\end{table}

\noindent\textbf{Physics-grounded object placement.}
\label{app:ablation-placement}
Table~\ref{tab:app-ablation-placement} compares raw SAM3D poses, the same poses after 5\,s of PhysX settling, and \method's combined initialization and pose-refinement pipeline. All variants share segmentation, assets, camera, root surfaces, materials, and lighting. The SAM3D variants retain the shared gravity-aligned frame but bypass our metric pose initialization and refinement, gravity-based stability checks, and physical settling during reconstruction. 
The post-hoc variant applies its 5\,s settling pass only after the raw-pose scene is assembled.
Raw poses often violate support relations: 69\% of objects lie below their support surface (median 17\,mm), and 15\% float above it. Only 2 of 80 scenes keep every object within 10\,mm of its initial position after settling, and 46 of 80 within the 50\,mm stability threshold. 
Post-hoc settling allows objects to reach resting configurations but can reduce segmentation overlap and 3D placement accuracy.
\method improves both alignment and placement while all 80 scenes pass the stability test. This comparison evaluates the combined placement pipeline, rather than isolating initialization and refinement individually.
To assess the bounded move tool, the code-edits-only variant removes it from pose refinement while retaining the scene-code editing tool, which settles and renders each edit for agent review. It obtains lower instance overlap (ARI 0.858 vs.\ 0.878; FG-IoU 0.792 vs.\ 0.819) and object appearance (LPIPS$^\ast$ 0.178 vs.\ 0.170), while all 80 scenes remain stable. The replay ablation reveals a larger downstream gap: success falls from 80\% to 44\% without the bounded move tool (Table~\ref{tab:replay-placement-ablation}). This suggests that aggregate reconstruction metrics can miss placement differences that matter for executing recorded robot motions.


\begin{table}[t]
    \centering
    \small
    \caption{\textbf{Object placement ablation.} SAM3D's raw poses share every other component (segmentation, assets, camera, root surfaces) with \method. ``Code edits only'' removes \texttt{Move}; the agent specifies pose changes through \texttt{Edit}, which still runs physical simulation on the edited scene.}
    \label{tab:app-ablation-placement}
    \vspace{-2mm}
    \setlength{\tabcolsep}{3pt}
    \resizebox{\linewidth}{!}{%
    \begin{tabular}{l|ccc|ccc|cc|cc}
        \toprule
        & \multicolumn{3}{c}{Source-view reconstruction}
        & \multicolumn{3}{c}{Object-centric appearance}
        & \multicolumn{2}{c}{Instance structure}
        & \multicolumn{2}{c}{3D object layout}\\
        \cmidrule(lr){2-4} \cmidrule(lr){5-7} \cmidrule(lr){8-9} \cmidrule(lr){10-11}
        Placement & PSNR $\uparrow$
        & LPIPS $\downarrow$ & nCLIP $\downarrow$
        & PSNR$^\ast$ $\uparrow$
        & LPIPS$^\ast$ $\downarrow$ & nCLIP$^\ast$ $\downarrow$
        & ARI $\uparrow$ & FG-IoU $\uparrow$
        & OSPA@5 $\downarrow$ & F1@10 $\uparrow$ \\
        \midrule
        SAM3D raw poses & 12.915 & 0.538 & 0.131
        & 12.810 & 0.195 & 0.078 & 0.784 & 0.705 & 2.7 & 0.88 \\
        SAM3D raw poses + physics settle & 12.768 & 0.546 & 0.129
        & 12.408 & 0.198 & 0.079 & 0.748 & 0.664 & 3.0 & 0.84 \\
        \method, code edits only & 13.272 & 0.523 & 0.124
        & 13.415 & 0.178 & 0.066 & 0.858 & 0.792 & \textbf{2.4} & 0.92 \\
        \method & \textbf{13.277} & \textbf{0.518} & \textbf{0.122}
        & \textbf{13.609} & \textbf{0.170} & \textbf{0.062}
        & \textbf{0.878} & \textbf{0.819} & \textbf{2.4} & \textbf{0.93}\\
        \bottomrule
    \end{tabular}}
\end{table}

\begin{table}[t]
    \centering
    \small
    \caption{\textbf{Reconstruction with different harnesses and shared VLM backbones.} We compare VIGA, \method, and the direct Codex harness.  VIGA and Codex appearance and layout metrics are computed after an additional 5\,s settling pass. Note that results from Tables~\ref{tab:vlm}~and~\ref{tab:layout3d} evaluate each method’s native reconstruction output. Stability is measured over the settling pass. API cost includes both VLM and image generation model costs.}
    \vspace{-2mm}
    \label{tab:app-astra-harness}
    \setlength{\tabcolsep}{3pt}
    \resizebox{\linewidth}{!}{%
    \begin{tabular}{llccc|ccc|cc|cc|c|c}
        \toprule
        & & \multicolumn{3}{c}{Full-scene appearance}
        & \multicolumn{3}{c}{Object-centric appearance}
        & \multicolumn{2}{c}{Instance structure}
        & \multicolumn{2}{c}{3D object layout} & Physics & API cost \\
        \cmidrule(lr){3-5} \cmidrule(lr){6-8} \cmidrule(lr){9-10} \cmidrule(lr){11-12} \cmidrule(lr){13-13} \cmidrule(lr){14-14}
        VLM backbone & \multicolumn{1}{|l|}{Method} & PSNR $\uparrow$
        & LPIPS $\downarrow$ & nCLIP $\downarrow$
        & PSNR$^\ast$ $\uparrow$ & LPIPS$^\ast$ $\downarrow$ & nCLIP$^\ast$ $\downarrow$
        & ARI $\uparrow$ & FG-IoU $\uparrow$
        & OSPA@5 $\downarrow$ & F1@10 $\uparrow$
        & Stable scenes $\uparrow$ & (\$/scene) $\downarrow$ \\
        \midrule
        \multirow{2}{*}{Claude Opus 5} & \multicolumn{1}{|l|}{VIGA} & 11.042 & 0.671 & 0.191
        & 8.725 & 0.337 & 0.154 & 0.261 & 0.239 & 4.9 & 0.20 & 8/80 & 16.13 \\
        & \multicolumn{1}{|l|}{\method} & \textbf{13.277} & \textbf{0.518} & \textbf{0.122}
        & \textbf{13.609} & \textbf{0.170} & \textbf{0.062}
        & \textbf{0.878} & \textbf{0.819} & \textbf{2.4} & \textbf{0.93}
        & \textbf{80/80} & \textbf{4.22} \\
        \midrule
        \multirow{3}{*}{GPT-6 Astra} & \multicolumn{1}{|l|}{VIGA} & 15.316 & 0.433 & 0.125
        & 14.077 & 0.187 & 0.082 & 0.721 & 0.648 & 4.6 & 0.37 & 24/80 & 7.74 \\
        & \multicolumn{1}{|l|}{Codex harness} & \textbf{17.007} & 0.370 & \textbf{0.100} & \textbf{14.773} & 0.178 & 0.073 & 0.653 & 0.594 & 4.2 & 0.60 & 16/80 & \textbf{3.99} \\
        & \multicolumn{1}{|l|}{\method} & 16.512 & \textbf{0.369} & \textbf{0.100} & 14.535 & \textbf{0.151} & \textbf{0.060} & \textbf{0.891} & \textbf{0.838} & \textbf{2.3} & \textbf{0.95} & \textbf{80/80} & 6.24 \\
        \bottomrule
    \end{tabular}}
    \vspace{-2mm}
\end{table}

\begin{figure}[t]
    \centering
    \includegraphics[width=\linewidth]{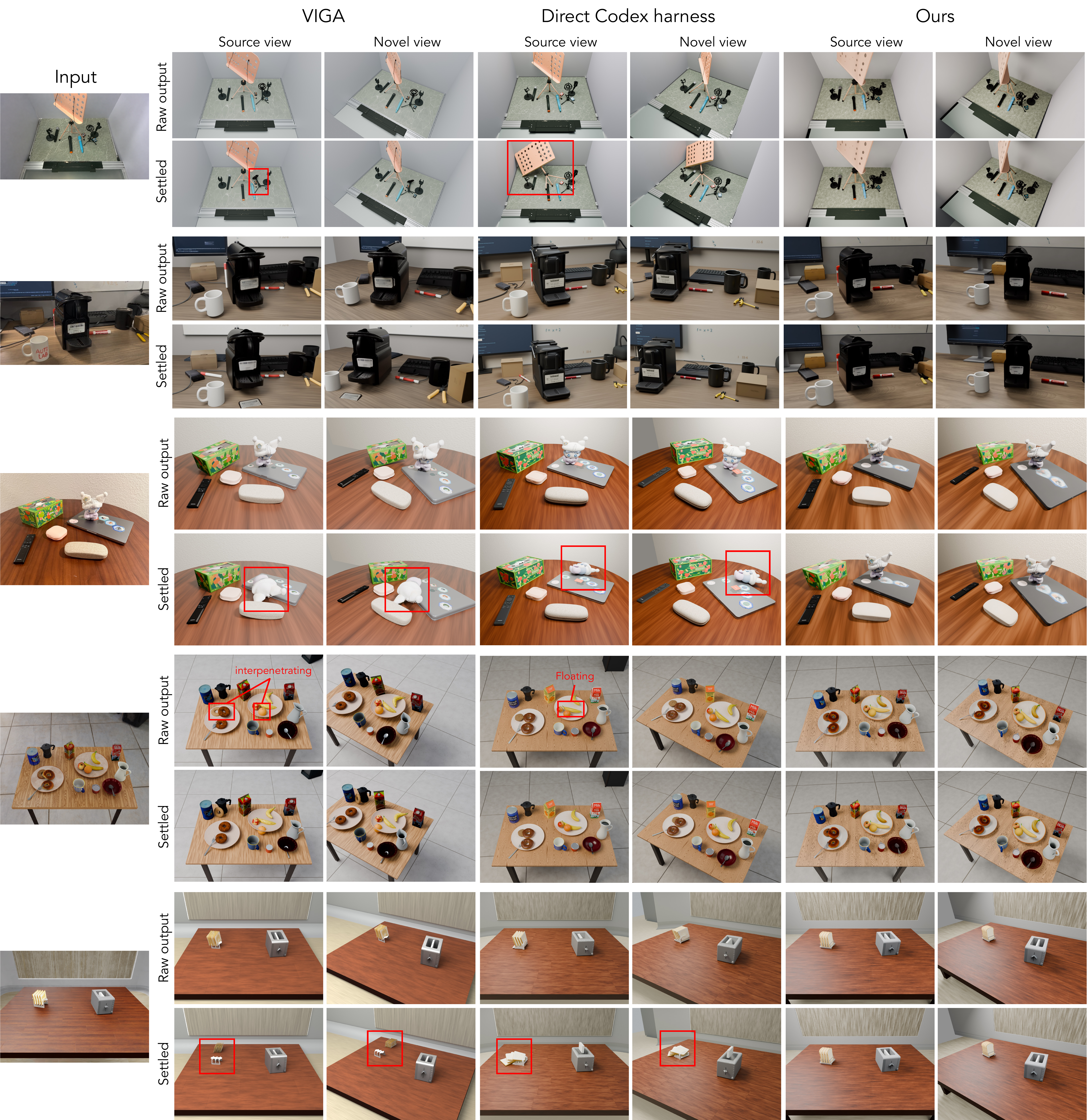}
    \vspace{-3mm}
    \caption{\textbf{Scene reconstruction with a shared GPT-6 Astra backbone.}}
    \label{fig:app-astra-harness}
    \vspace{-3mm}
\end{figure}

\subsection{Additional Quantitative Results}
\label{app:additional-quantitative}

\noindent\textbf{Direct Codex harness with GPT-6 Astra.}
\label{app:codex-direct}
Codex~\citep{openai2025codex} is OpenAI's general-purpose coding-agent meta-harness: it gives the model shell and file tools but no task-specific structure.
For the direct Codex baseline in Table~\ref{tab:app-astra-harness}, we attach
each benchmark image to an isolated GPT-6 Astra session with \texttt{ultra}
reasoning effort.  We automate the runs through the non-interactive
\texttt{codex exec} interface, using the prompt shown in Listing~\ref{lst:codex-direct-prompt}.  The agent writes Blender Python, runs Blender headlessly, inspects rendered images, and iteratively refines the camera, geometry, materials, lighting, and composition.  Each worker uses Blender 4.5.3 with Cycles on one assigned CUDA GPU, with a six-hour timeout per attempt and at most two attempts per image.  Filesystem isolation prevents access to other scene workspaces and reconstruction results.  The prompt prohibits external models, textures, images, and assets, and requires a final-render inspection followed by at least one refinement pass.  It also requires one semantically labeled top-level Empty per independently movable object, with all of that object's parts grouped beneath it, so that each object can be simulated as a single rigid body; environment geometry, cameras, and lights have separate roots.  Required outputs are a procedural \texttt{scene.py}, a self-contained \texttt{final.blend}, its rendered \texttt{final.png}, and a \texttt{report.json} describing reconstruction decisions and known mismatches.

\noindent\textbf{Reconstruction harnesses with shared VLM backbones.}
Table~\ref{tab:app-astra-harness} compares reconstruction harnesses with a
shared VLM backbone. Both VIGA and the direct Codex harness receive an
object-grouping instruction: all parts of each movable object share a parent
Empty, allowing it to be simulated as a single rigid body. We run VIGA
with medium reasoning effort and a 20-round budget for both backbones.
With Claude Opus 5, \method outperforms VIGA
across appearance, instance structure, object placement, and physical
stability.  With GPT-6 Astra we compare against two baselines that share the
backbone but not our harness.  Neither VIGA
nor the direct Codex harness has a physics stage, so we score their scenes
after a 5\,s PhysX settle with every object as a rigid body with its parts (primitives) being convex-hull
colliders, letting floating and interpenetrating objects come to rest before
the metrics are computed.  Astra substantially improves VIGA's
appearance, and the Codex harness achieves the highest full-scene PSNR.
\method nevertheless achieves better instance segmentation, more accurate
placement, and a higher stability pass rate: all 80 scenes pass, compared with
24 for VIGA and 16 for Codex, while OSPA@5 is 2.3\,cm compared with
4.6 and 4.2\,cm, respectively. These results indicate that improved image-space
fidelity alone does not ensure accurate and stable object placement.
\method complements VLM reasoning with explicit scene structure, metric
grounding, and physics-checked pose corrections; this comparison evaluates
the complete harnesses rather than individual components.
Figure~\ref{fig:app-astra-harness} illustrates the comparison.

\begin{lstlisting}[language={},basicstyle=\ttfamily\scriptsize,
    columns=fullflexible,keepspaces=true,showstringspaces=false,
    caption={Task prompt for the direct Codex baseline.},
    label={lst:codex-direct-prompt}]
Reconstruct the attached target image as a procedural Blender scene.

The same reference image is available at `{{TARGET_FILE}}` inside your current
workspace. Work autonomously
and iteratively: write Blender Python, run Blender headlessly, render candidate
images, inspect those renders with the image-viewing tool, and improve the
camera, geometry, materials, lighting, and composition until the result is a
strong visual match.

Constraints:

- Work only inside the current workspace; do not search for other files or solutions.
- Do not download or use external models, textures, images, or other assets.
- Blender and Blender's embedded Python are the only application/runtime you
  should use for scene creation and rendering.
- Prefer a procedural, reproducible `scene.py` over manual-only edits.
- Use headless Blender and render with Cycles on the assigned CUDA GPU. CPU
  rendering is not an acceptable fallback in this benchmark. In every script
  that renders, initialize Cycles with the Blender 4.5-compatible sequence:

  ```python
  scene.render.engine = "CYCLES"
  preferences = bpy.context.preferences.addons["cycles"].preferences
  preferences.compute_device_type = "CUDA"
  preferences.refresh_devices()
  cuda_devices = [device for device in preferences.devices if device.type == "CUDA"]
  if not cuda_devices:
      raise RuntimeError("Assigned CUDA GPU is unavailable")
  for device in preferences.devices:
      device.use = device.type == "CUDA"
  scene.cycles.device = "GPU"
  ```

  Do not use `get_devices()` in place of `refresh_devices()`, and do not catch
  CUDA initialization errors by falling back to CPU. Confirm the final render
  log identifies a CUDA device and record `Cycles CUDA` in `report.json`.
- Match the reference image's framing and aspect ratio.
- The final `.blend` must render without external file dependencies.
- Do not stop at an explanation or plan; create and validate the artifacts.

Object grouping requirement:

- Create one top-level Empty for each independently movable physical object.
- Parent every mesh, curve, text object, and decorative part belonging to that
  object under its Empty.
- If several parts would normally move together as one rigid object, they
  belong to the same Empty. For example, a mug body, handle, liquid, and logo
  should share one parent.
- Do not combine separate physical objects under one Empty, even when they
  touch or overlap. Do not create a separate Empty for every mesh part.
- Name object roots `OBJ_<semantic_label>_<index>`, such as `OBJ_mug_01`, and
  set these custom properties on each root:

  ```python
  root["is_object_asset"] = True
  root["semantic_label"] = "mug"
  ```

- Put non-object geometry under the reserved top-level Empty `ENVIRONMENT`.
- Put cameras and lights under `CAMERA_RIG` and `LIGHTING_RIG`.
- Only Empty objects may exist at the top level. Every renderable object must
  ultimately descend from one of these top-level groups.

Before saving `final.blend`, validate the grouping:

```python
renderable_types = {"MESH", "CURVE", "FONT", "SURFACE", "META"}

top_level = [obj for obj in bpy.context.scene.objects if obj.parent is None]
assert all(obj.type == "EMPTY" for obj in top_level), [
    obj.name for obj in top_level if obj.type != "EMPTY"
]

for obj in bpy.context.scene.objects:
    if obj.type not in renderable_types:
        continue
    root = obj
    while root.parent is not None:
        root = root.parent
    assert root.type == "EMPTY", obj.name
    assert (
        root.get("is_object_asset", False)
        or root.name in {"ENVIRONMENT", "CAMERA_RIG", "LIGHTING_RIG"}
    ), f"{obj.name} is not assigned to a recognized group"
```

Required deliverables:

- `output/scene.py`
- `output/final.blend`
- `output/final.png`
- `output/report.json`

`report.json` must contain the scene id `{{SCENE_ID}}`, the chosen renderer,
image dimensions, the main reconstruction decisions, and any known mismatch.
Before finishing, open `output/final.png` with the image-viewing tool
and make at least one refinement pass based on that inspection.
\end{lstlisting}

\noindent\textbf{Segmentation before and after 3D reconstruction.}
Table~\ref{tab:app-stage1-seg} scores each method's own 2D instance segmentation of the input and the instance masks rendered from its final scene, against the simulator ground truth of the 44 grounded scenes.
\method achieves high input-mask agreement with the ground truth (ARI 0.981 and FG-IoU 0.970). In contrast, although REST3D and SceneConductor also segment well, their results degrade notably once objects are placed in 3D. The decline from input segmentation to rendered instance masks highlights errors introduced during lifting and reconstruction, including geometry, camera alignment, scale, and object placement.

\begin{table}[!htb]
    \centering
    \small
    \caption{\textbf{Instance segmentation at the segmentation stage and in the final scene.}
    Results on the 44 simulator-grounded scenes. Stage-1 metrics use visible instance masks on the original input image, whereas final-scene counts include only objects visible in the source-view render. Changes in visibility can therefore change the count even without a change in the reconstructed object inventory.}
    \vspace{-2mm}
    \label{tab:app-stage1-seg}
    \resizebox{0.8\linewidth}{!}{%
    \begin{tabular}{l|ccc|ccc}
        \toprule
        & \multicolumn{3}{c}{Stage-1 segmentation} & \multicolumn{3}{c}{Final scene} \\
        \cmidrule(lr){2-4} \cmidrule(lr){5-7}
        Method & ARI $\uparrow$ & FG-IoU $\uparrow$ & Count MAE $\downarrow$
        & ARI $\uparrow$ & FG-IoU $\uparrow$ & Count MAE $\downarrow$ \\
        \midrule
        VIGA & 0.450 & 0.394 & 4.43 & 0.283 & 0.260 & 3.36 \\
        REST3D & \underline{0.864} & \underline{0.809} & 3.18 & 0.210 & 0.204 & 3.20 \\
        SceneConductor & 0.824 & 0.800 & \underline{1.11} & 0.197 & 0.187 & 1.14 \\
        SimFoundry & 0.780 & 0.732 & 1.84 & 0.676 & 0.589 & \underline{1.05} \\
        \method (ours) & \textbf{0.981} & \textbf{0.970} & \textbf{0.27} & \textbf{0.878} & \textbf{0.819} & \textbf{0.27} \\
        \bottomrule
    \end{tabular}
    }
\end{table}

\subsection{Additional Qualitative Results}
\label{app:additional-qualitative}

\begin{figure}[h]
    \centering
    \includegraphics[width=\linewidth]{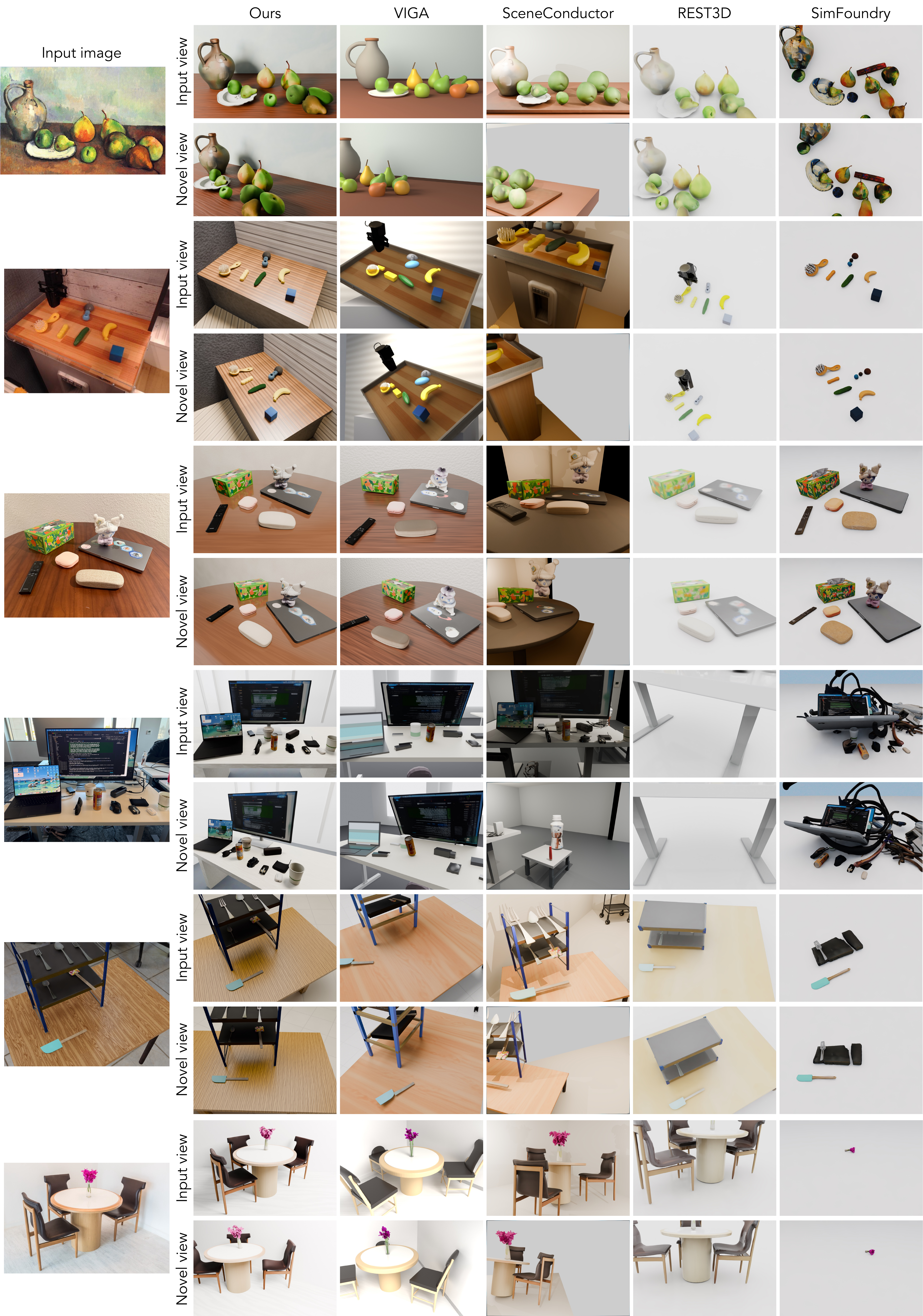}
    \caption{\textbf{Additional qualitative reconstruction results.}}
    \label{fig:additional-qualitative}
\end{figure}

Figure~\ref{fig:additional-qualitative} extends the comparison to more scenes.
Across these examples, \method more closely preserves object arrangement,
small items, and surrounding surfaces.  The novel views reveal missing geometry
and incorrect depth placement that can be less apparent from the input
viewpoint, highlighting the importance of reconstructing individual objects
and their spatial relationships.

\end{document}